\def\paperlanguage{} 

\documentclass[letterpaper, 10 pt, conference]{ieeeconf}  

\usepackage{bm}
\usepackage{cite}
\usepackage{flushend}
\include{preamble}

\usepackage{booktabs}

\IEEEoverridecommandlockouts                              

\title{\LARGE \textbf
  {
    \switchlanguage%
    {%
      Patterned Structure Muscle : Arbitrary Shaped Wire-driven 
      \\ Artificial Muscle 
      Utilizing Anisotropic Flexible Structure
      \\ for Musculoskeletal Robots
    }%
    {%
      3D-printed TPUによるワイヤ駆動の異方特性パターン構造筋
      for musculoskeletal robot
    }%
  }
}

\author{Shunnosuke Yoshimura$^{1}$, Akihiro Miki$^{1}$, Kazuhiro Miyama$^{1}$, Yuta Sahara$^{1}$,\\  Kento Kawaharazuka$^{1}$, Kei Okada$^{1}$, and Masayuki Inaba$^{1}$
  \thanks{$^{1}$ The authors are with the Department of Mechano-Informatics, Graduate School of Information Science and Technology, The University of Tokyo, 7-3-1 Hongo, Bunkyo-ku, Tokyo, 113-8656, Japan.
    {\texttt\small [yoshimura, miki, miyama, sahara, kawaharazuka, okada, inaba]@jsk.t.u-tokyo.ac.jp}
  }
}
\begin{document}

\maketitle
\thispagestyle{empty}
\pagestyle{empty}

\begin{abstract}
  \switchlanguage%
  {%
  Muscles of the human body are composed of tiny actuators made up of myosin and actin filaments. 
  They can exert force in various shapes such as curved or flat, 
  under contact forces and deformations from the environment.
  On the other hand, muscles in musculoskeletal robots so far have faced challenges in 
  generating force in such shapes and environments.
  To address this issue, we propose Patterned Structure Muscle (PSM), 
  artificial muscles for musculoskeletal robots.
  PSM utilizes
  patterned structures with anisotropic characteristics, wire-driven mechanisms,
  and is made of flexible material Thermoplastic Polyurethane (TPU) using FDM 3D printing.
  This method enables the creation of various shapes of muscles,
  such as simple 1 degree-of-freedom (DOF) muscles, 
  Multi-DOF wide area muscles, joint-covering muscles, and branched muscles.
  We created an upper arm structure using these muscles to demonstrate
  wide range of motion, lifting heavy objects, 
  and movements through environmental contact.
  These experiments show that the proposed PSM is capable of 
  operating in various shapes and environments,
  and is suitable for the muscles of musculoskeletal robots.
  }%
  {%
  人間の筋は、ミオシン・アクチンフィラメントからなる微小なアクチュエータが集積して構成される。
  ゆえに、曲面形状から広い面形状など、
  多様な形状で、かつ環境からの接触力や変形のもとで、柔軟に動作することができる。
  一方、これまでの筋骨格ロボットの筋は、そのような形状や環境下での動作が困難であった。
  そこで我々は、弾性および可動域に異方特性を有する構造のパターン化の活用および、
  柔軟素材であるTPUの3Dプリント技術を用いることで、
  ワイヤ駆動メカニズムとしてこの筋の性質を実現した
  Patterned Structure Muscle (PSM)を作成した。
  この手法は、
  積層3Dプリンタによる容易な製造が可能でありながら、
  単純な1筋・広い面状の筋・関節を覆う筋・分岐した筋といった
  様々な形状の筋を作成可能である。
  さらに、これらの筋を組み合わせた上腕構造を作成し、
  広い可動範囲の動作・重量物持ち上げ・物体と筋の接触下での
  動作を行った。
  これらの実験により、提案したPSMが、様々な形状・環境下での動作が可能であり、
  筋骨格ロボットの筋として利用可能であることを示した。
  }%
\end{abstract}

\section{Introduction}\label{sec:introduction}
\begin{figure}[t]
  \centering
  \includegraphics[width=1.0\columnwidth]{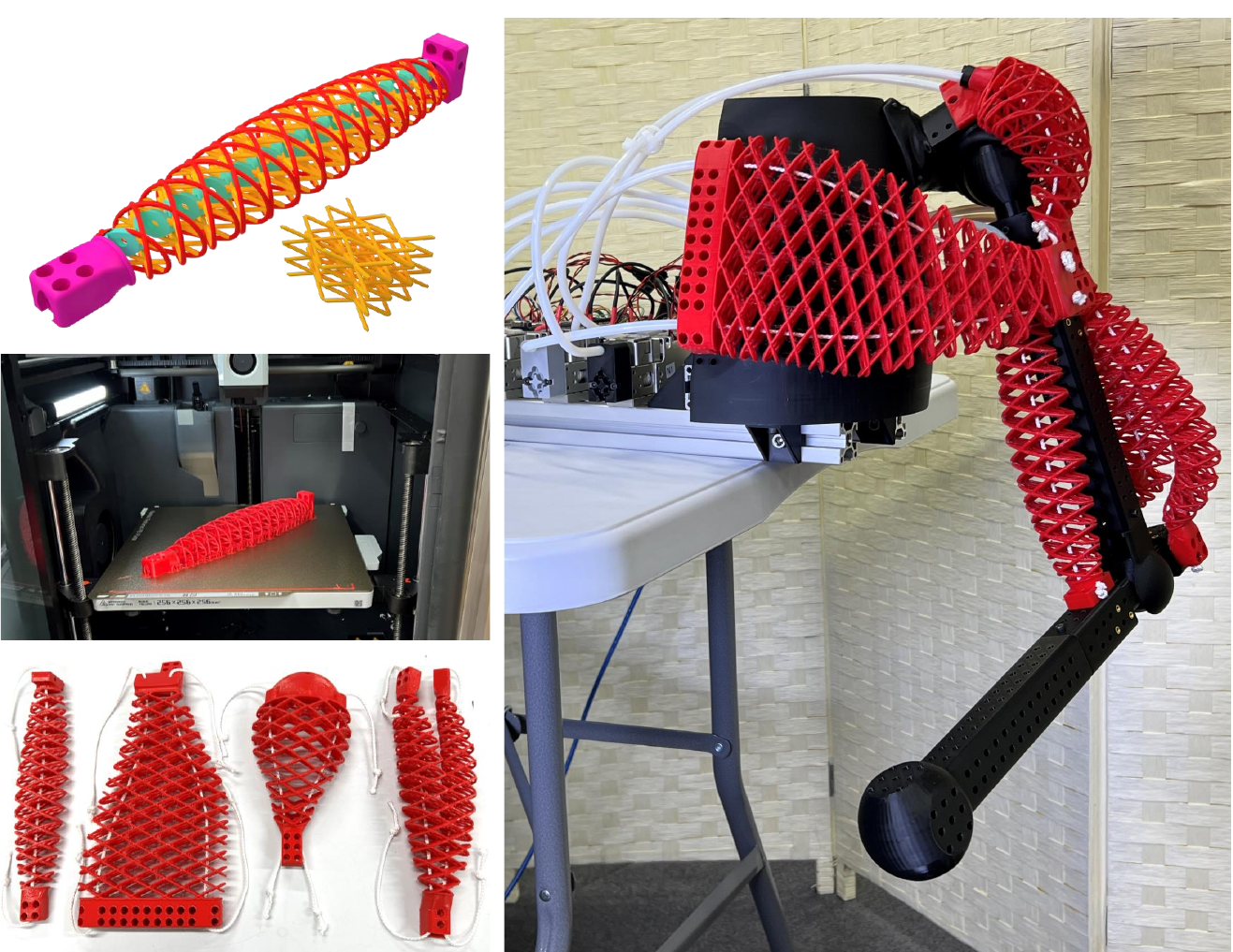}
  \vspace{-3ex}
  \caption{
    The overview of this study: On the left, the structure of PSM, 
    the process of making PSM using 3D printing, 
    and the four types of PSM are shown. 
    On the right, the upper arm structure driven by PSM is shown. 
  }
  \label{figure:overview}
  \vspace{-4ex}
\end{figure}

\switchlanguage%
{%
Human muscles generate contraction movements through the exertion of force 
by myosin and actin filaments.
Each of the tiny actuators generates contraction force,
and the entire muscle contracts while maintaining the relative position of the actuators.
This mechanism allows muscles to operate in various shapes,
from thin muscles to wide area muscles and muscles distributed in curved shapes.
Moreover, muscles can operate under contact forces from environment 
and deformations from other muscles' movements flexibly.
Imitating this mechanism of moving the body through muscles and implementing musculoskeletal robots 
is important, for the purpose of creating robots closer to humans 
and for deepening our understanding of human structure and movement.
Indeed, in the history of musculoskeletal robots, 
muscle actuators have been implemented using methods such as wire-driven systems 
and pneumatic systems
\cite{asano2016kengoro, Musashi, marques2010ecce1, mouthuy2022humanoid,
kurumaya2016musculoskeletal, IkemotoShoulder, Niiyama2010_Humanoids2010_AthleteRobot}.
However, these methods have not been able to realize the various shapes of human muscles, 
and operation under contact between muscles and the environment.
On the other hand, in recent years, 
advancements in soft robotics technology and 3D printing have progressed, 
enabling the application of flexible mechanisms to various fields.
Therefore, we propose artificial muscles for a musculoskeletal robot, 
called Patterned Structure Muscle (PSM),
by combining wire-driven methods with soft robotics and 3D printing technologies.
The overview of this research is shown in \figref{figure:overview}.
In this method, the mechanism of the contraction movement of the human muscle is imitated
by combining a flexible and deformable 3D pattern structure and a wire-driven mechanism.
This structure has anisotropic characteristics and a wide range of motion,
which allows it to operate under contact forces and flexible deformations.
We create various shapes of muscles using this method,
such as simple 1 degree-of-freedom (DOF) muscles,
multi-DOF wide area muscles, joint-covering muscles, and branched muscles.
To demonstrate the practicality of PSM and its suitability for musculoskeletal robots, 
we conduct three experiments.
Firstly, we create muscles with various parameters, and measure their characteristics.
Secondly, we confirm that the muscles of various shapes can operate individually in two different situations.
Finally, we create an upper arm structure using these muscles and demonstrate its wide range of motion,
lifting heavy objects, and movements through contact between the muscles and the environment.
}%
{%
人間の筋は、ミオシン・アクチンフィラメントが力を発揮することで収縮運動を生み出す。
それぞれの小さなアクチュエータが収縮力を発揮することにより、アクチュエータ同士の位置関係を
維持しながら、筋全体として収縮変形が生じる。
小さいアクチュエータが集積して動作するこのメカニズムは、
細い筋から広い面状の筋、曲面状に分布する筋まで、多様な形状で動作する。
さらに、柔軟性を有しつつも、環境からの接触力や、
他の筋の駆動による関節角度変化といった変形のもとで動作可能である。
このような、筋により身体を動作させる仕組みを模倣し、筋骨格ロボットを実現することは、
人間により近いロボットを実現する目的でも、
人間の構造や動作の理解を深める目的でも、重要である。
実際、これまでの筋骨格ロボットにおいても、ワイヤ駆動や空気圧を活用して
筋アクチュエータが実現されてきた
\cite{asano2016kengoro, Musashi, marques2010ecce1, mouthuy2022humanoid,
kurumaya2016musculoskeletal, IkemotoShoulder, Niiyama2010_Humanoids2010_AthleteRobot}。
しかし、人間の筋に相応する多種類の形状での動作や、接触動作の実現はできていない。
一方、近年のソフトロボット技術、3Dプリント技術の進展により、柔軟性を有する
応用可能なメカニズムの開発が進んでいる。
そこで我々は、弾性および可動域に異方特性を有するパターン構造の活用および、
柔軟素材であるTPUの3Dプリント技術を用いて、
筋骨格ロボットの筋アクチュエータとして可能なワイヤ駆動メカニズム、
Patterned Structure Muscle(PSM)を作成した。
この概要を\figref{figure:overview}に示す。
この手法では、人間の筋において微小アクチュエータ同士が位置関係を保ちながら変形するメカニズムを、
柔軟に変形可能な3次元パターン構造とワイヤ駆動の組み合わせとして再現している。
また、可動域・弾性に異方性を有する構造により、筋の柔らかさと、環境接触・変形下での動作を両立している。
そこで、PSMの基本的構造と特性を示した上で、
単純な1筋・広い面状の筋・関節を覆う筋・分岐した筋を作成し、それぞれが動作することを確認する。
さらに、これらの筋を組み合わせた上腕構造を作成し、広い可動範囲の動作・重量物持ち上げ・
物体と筋の接触下での動作を行う。
これらの実験により、
提案したPSMが、人間の筋のような形状・動作を有し、筋骨格ロボットのアクチュエータに適していることを示す。
}%
\subsection{Related Work}
\switchlanguage%
{%
In recent years, musculoskeletal robots have been developed, 
which mimic the human musculoskeletal structure.
Kengoro\cite{asano2016kengoro} and Musashi\cite{Musashi} have more than 70 muscles driven by wires,
and can perform human-like movements by combining a skeletal structure and muscles.
ECCE1\cite{marques2010ecce1} has a wire-driven upper body structure,
and Roboy\cite{mouthuy2022humanoid} combines wire-driven and biological tendons.
The lower-limb robot\cite{kurumaya2016musculoskeletal} developed by Kurumaya et al.
uses multifilament McKibben muscles to perform simple leg movements.
Ikemoto et al. have developed a shoulder complex using pneumatic actuators\cite{IkemotoShoulder}.
The Athlete Robot\cite{Niiyama2010_Humanoids2010_AthleteRobot} also uses 
pneumatic actuators to perform dynamic movements.
Thus, the actuators of musculoskeletal robots have been mainly wire-driven muscles
\cite{asano2016kengoro, Musashi, marques2010ecce1, mouthuy2022humanoid}
and pneumatic artificial muscles
\cite{kurumaya2016musculoskeletal, IkemotoShoulder, Niiyama2010_Humanoids2010_AthleteRobot}.
Wire-driven muscles have the advantage of being able to control 
the length and tension of the wire accurately,
and are not dependent on the length and force of the wire so that 
they can be used for long contraction distances.
However, they have the disadvantage of being difficult 
to perform contact operations with the environment,
and are difficult to operate in wide and curved shapes.
Pneumatic artificial muscles have the advantage of being able to
exert large forces even under environmental contact,
but have the disadvantage of having a limited range of motion,
and there are constraints on the possible shapes of muscles that can be created.

On the other hand, the development of soft actuators using flexible materials has been advanced, 
which partially overcomes these weaknesses.
Elastomeric Origami\cite{https://doi.org/10.1002/adfm.201102978},
Origami structures of various shapes\cite{https://doi.org/10.1002/advs.201901371},
and 3D printed soft actuators\cite{schaffner20183d},
have been developed using elastomer materials with high stretchability and large deformability,
but have the disadvantage of being difficult to perform contact operations and 
to be used for large muscles.
HASEL actuator\cite{doi:10.1126/science.aao6139} and Peano-HASEL actuators\cite{doi:10.1126/scirobotics.aar3276}
are electrohydraulic actuators using flexible materials, and have been applied to circular muscles\cite{https://doi.org/10.1002/adfm.201908821},
but it is difficult to use them for large three-dimensional muscles.
Hybrid carbon nanotube yarn muscles\cite{doi:10.1126/science.1226762} and
artificial muscles from fishing line and sewing thread\cite{doi:10.1126/science.1246906}
are muscles made of threads, and have the advantage of being durable, but 
they do not have thickness, and are difficult to use for thick three-dimensional muscles.
Fluid-driven origami-inspired artificial muscles\cite{doi:10.1073/pnas.1713450114}
have compressible pattern structures and skins, and is able to exert large forces,
but is difficult to perform contact operations due to the thin skin.
Shape memory McKibben muscles\cite{TAKASHIMA2010116} and
high backdrivable McKibben muscles\cite{NA2023114381} have been developed,
but have constraints on the adaptable muscle shape.

Thus far, with soft actuators, various methods have been proposed to achieve flexibility, 
high deformability, and substantial force through different materials and mechanisms. 
However, when it comes to practical use in combination with the skeleton of a musculoskeletal robot, 
challenges remain regarding applications to muscles of various shapes 
and environmental contact movements.

Therefore, we aim to create muscles with the following features:
\begin{itemize}
  \item The proposed artificial muscle, PSM, combines soft robotics and wire-driven mechanisms, 
  allowing for flexibility while enabling large force generation and long-distance contraction.
  \item Similar to human muscles, PSM can be created in various shapes.
  \item PSM can operate under environmental contact and also safely,
  and can be used as muscles for musculoskeletal robots by combining multiple PSMs.
\end{itemize}
}%
{%
近年、人間の筋骨格構造を模倣した、筋骨格ロボットが作られてきた。
Kengoro\cite{asano2016kengoro}およびMusashi\cite{Musashi}は、
モータによるワイヤ駆動の筋を全身に100本以上搭載している。
背骨をはじめとする模倣的な関節構造と筋を組み合わせ、人間のような動作を行うことができる。
ECCE1をはじめとするECCERobot projectのロボット\cite{marques2010ecce1}もまた、
人間の上半身を模倣したワイヤ駆動ロボットであるほか、
Roboy\cite{mouthuy2022humanoid}では、
ワイヤ駆動と生体靭帯を組み合わせている。
Kuruyamaらが開発したlower-limb robot\cite{kurumaya2016musculoskeletal}
では、multifilament McKibben musclesを活用し、
模倣的な筋骨格脚を吊り下げて、単純な動作を行わせた。
Ikemotoらは、空気圧アクチュエータにより、肩複合体から上腕、手までを動作させている\cite{IkemotoShoulder}。
空気圧アクチュエータにより動的な動作を実現した
アスリートロボット\cite{Niiyama2010_Humanoids2010_AthleteRobot}も存在する。
このように、これまでの筋骨格ロボットのアクチュエータは、ワイヤ駆動により筋を再現したもの
\cite{asano2016kengoro, Musashi, marques2010ecce1, mouthuy2022humanoid}
および、
空気圧アクチュエータにより筋を再現したもの
\cite{kurumaya2016musculoskeletal, IkemotoShoulder, Niiyama2010_Humanoids2010_AthleteRobot}
が主流である。
ワイヤ駆動は、長い収縮距離、長さと力の大きさに依存性がない、
長さや張力の制御が容易であるといった利点がある。
しかし、ワイヤが体外に露出し身体と環境の接触動作が困難であること、
ワイヤのたるみや絡まりによる動作不良、広い面状や曲面形状での動作が困難であることといった
課題がある。
また、空気圧アクチュエータにおいては、McKibben musclesのように接触動作を得意とする種類も存在するが、
可動域が限られること、筋の形状に制約があることといった課題がある。

一方、これらの弱点を部分的に克服するような、柔軟素材を活用したソフトアクチュエータの開発も進んでいる。
Elastomeric Origami\cite{https://doi.org/10.1002/adfm.201102978}や、
様々な形状のOrigami構造の利用\cite{https://doi.org/10.1002/advs.201901371}, 
3Dプリントによるソフトアクチュエータ\cite{schaffner20183d}
では、
エラストマー素材の高い伸縮性および大変形に適した構造を空気圧等と組み合わせて駆動するが、
大きな筋への活用や、接触動作に課題がある。
筋肉に近い性能を有するHASEL actuator\cite{doi:10.1126/science.aao6139}および、
俊敏で大きな力を出せるPeano-HASEL actuators\cite{doi:10.1126/scirobotics.aar3276}は、
柔軟素材を使用したelectrohydraulic actuatorであり、
生物模倣的なCircular muscleへの応用\cite{https://doi.org/10.1002/adfm.201908821}も行われてきたが、
大きな三次元形状の筋への活用に課題がある。
Hybrid carbon nanotube yarn muscles\cite{doi:10.1126/science.1226762}や
Artificial muscles from fishing line and sewing thread\cite{doi:10.1126/science.1246906}は、
糸により筋肉を構成することで、耐久性が高く、長い筋肉への利用、大きな力と収縮率を実現しているが、
人間の筋にみられるような厚みのある三次元形状への活用に課題がある。
Fluid-driven origami-inspired artificial muscles\cite{doi:10.1073/pnas.1713450114}は、
compressible pattern strucutureとskin, fluid mediumを組み合わせることで、
大変形と大きな力密度を実現しているが、液体を薄いskinで覆う構造上、接触動作に課題がある。
形状記憶可能なMcKibben muscles\cite{TAKASHIMA2010116}や、
バックドライバビリティの高い
Braided thin McKibben muscles\cite{NA2023114381}も開発されてきたが、
適応可能な筋の形状に制約がある。
このように、これまでのソフトアクチュエータでは、
様々な素材・駆動方式により柔軟性や大変形性、大きな力を実現する手法が提案されてきたものの、
筋骨格ロボットの筋として、骨格と組み合わせて実際に使う上では、
形状の自由度や環境接触動作に課題が残されてきた。

そこで、我々が提案する手法では、次のような特徴を持つ筋を作成することを目指す。
\begin{itemize}
  \item ソフトロボットとワイヤ駆動を組み合わせ、
  柔軟性を有しながらも、大きな力・長い距離の収縮を可能にする。
  \item 人間の筋のように、様々な形状で製作できる。
  \item 環境接触下でも安全に動作し、複数の筋を組み合わせて筋骨格ロボットに利用できる。
\end{itemize}
}%


\section{Method} \label{sec:method}
\subsection{Concept of PSM} \label{subsec:concept}
\begin{figure}[t]
  \centering
  \includegraphics[width=1.0\columnwidth]{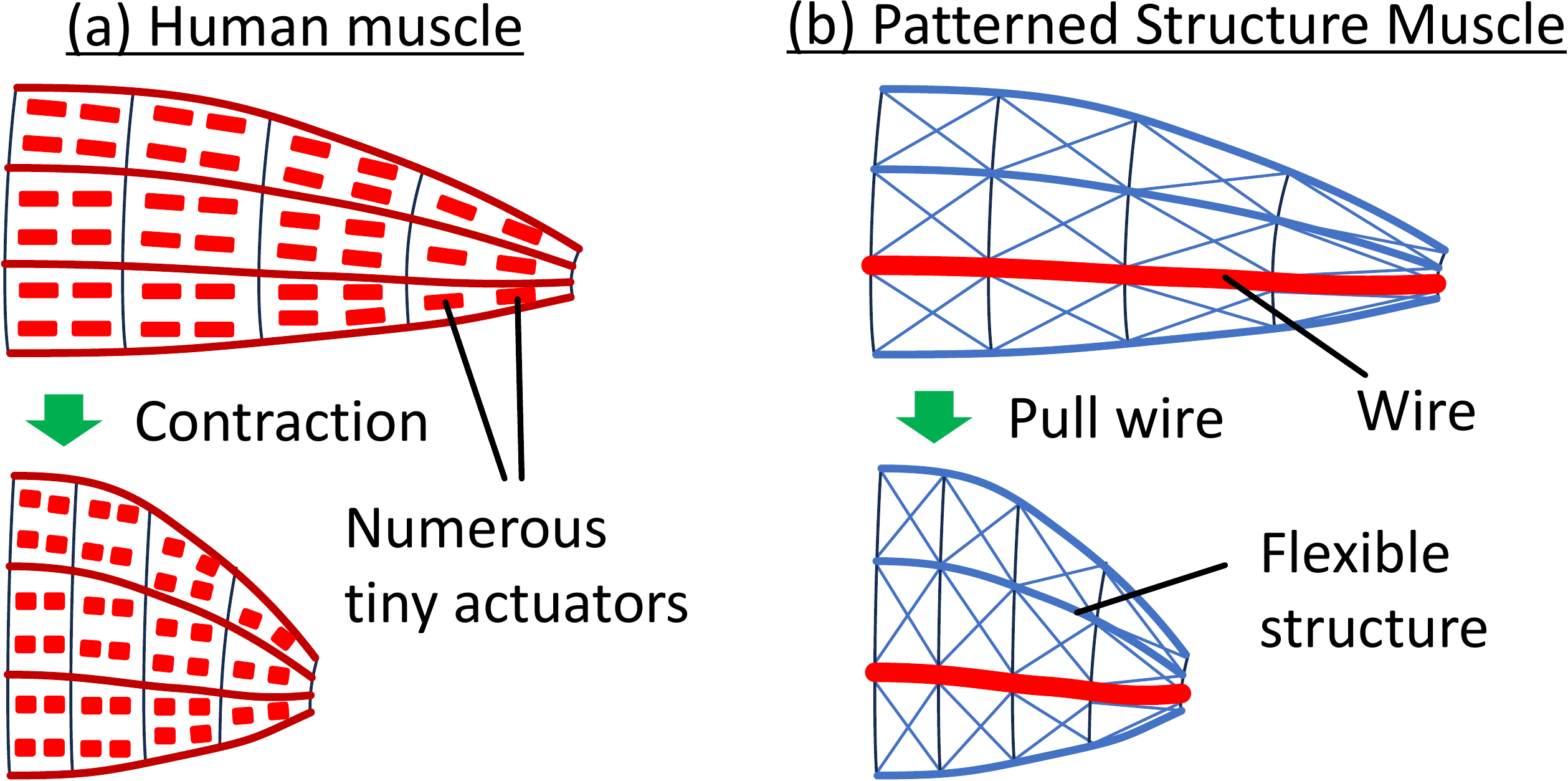}
  \caption{%
  Comparison between human muscle structure and the concept of PSM.
  While human muscle maintain their shape through the integration of small actuators
  during movement,
  PSM realizes similar movements through the combination of wire and flexible pattern structure.
  }
  \label{figure:concept}
  \vspace{-4ex}
\end{figure}

\begin{figure}[t]
  \centering
  \includegraphics[width=1.0\columnwidth]{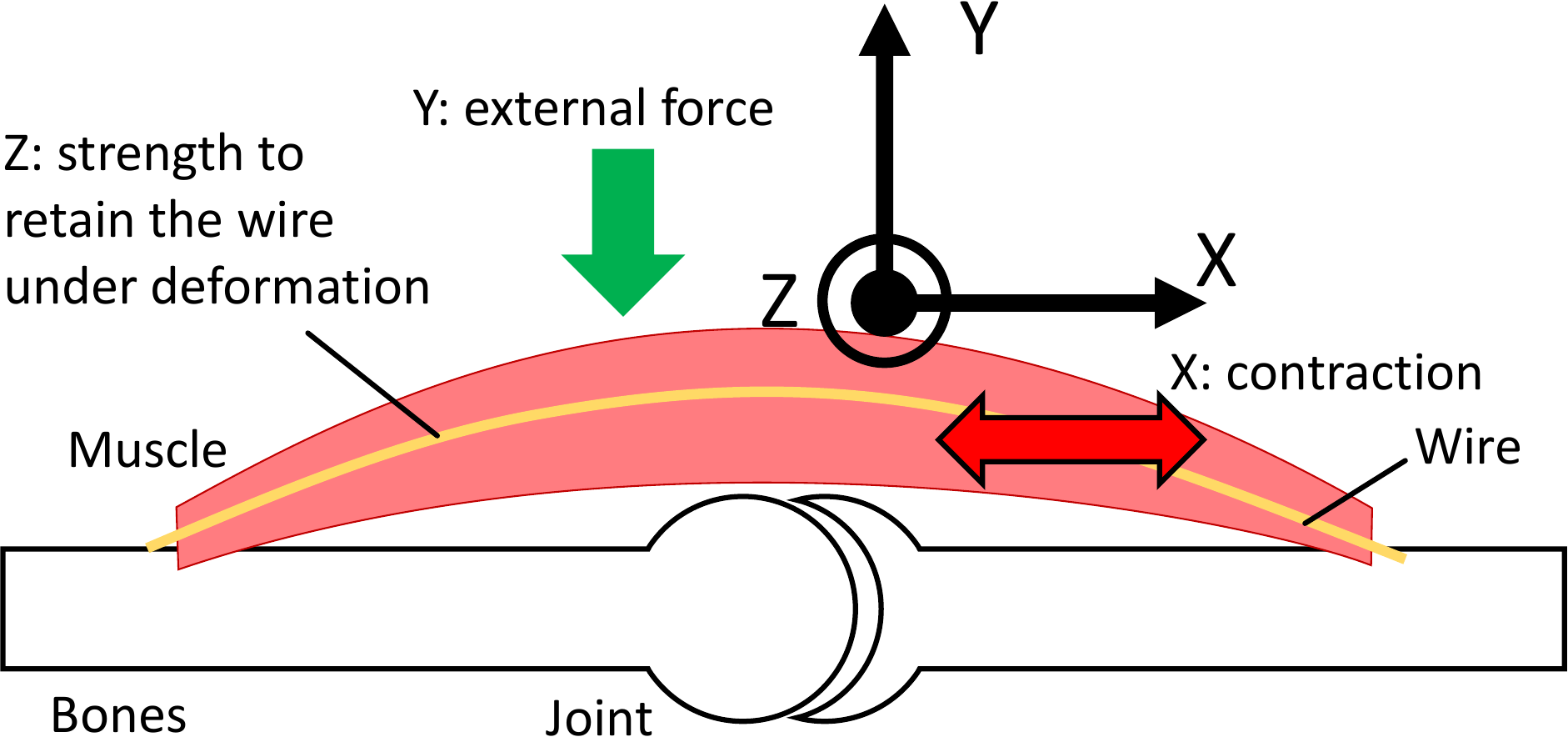}
  \vspace{-3ex}
  \caption{%
  The three directions of characteristics required for PSM.
  In x-direction, flexibility and range of motion are required 
  to realize contraction.
  In y-direction, strength is required to support the structure
  and allow x-directional contraction in the presence of external load.
  In z-direction, a certain degree of strength is required to maintain
  the positional relationship of wire and structure.
  }
  \label{figure:coord}
\end{figure}

\switchlanguage%
{%
Human muscle structure and the concept of PSM are compared in \figref{figure:concept}.
In human muscles, the myosin and actin filaments generate contraction movements,
causing myofibrils and muscle fibers to exert contraction force.
This mechanism, where small actuators drive while maintaining their relative positions in the muscle,
enables contraction even in situations such as deformation due to external forces.
In this study, we imitate this mechanism of the muscle, and realize it as a wire-driven mechanism.
In traditional wire-driven muscles, 
a wire is stretched between the end points of the muscle, 
and contraction is induced by winding the wire with a motor. 
However, because the wire itself lacks thickness, 
structures resembling various shapes of
human muscles such as sheet-like muscles 
cannot be achieved as they are. 
Additionally, since the wire is stretched linearly, 
performing actions with curved shapes is also challenging.
Therefore, we aim to address these issues through the following approach.
The wire is routed along the path of a single muscle fiber 
between the endpoints of the muscle. 
In areas without wire, flexible structure is created to 
deform while maintaining the shape of the muscle in accordance with the contraction of the wire. 
This allows for movements that are closer to human muscles.
}%
{%
人間の筋の構造と提案手法PSMの動作コンセプトの比較を\figref{figure:concept}に示す。
人間の筋は、ミオシンフィラメント・アクチンフィラメントの相互作用により、
筋原繊維が収縮する。沢山の筋原繊維が組み合わさった筋繊維、さらにそれらが組み合わさった筋は、
外力による変形などの状況下でも、
筋内部の相対位置をある程度保ちながら、全体として収縮変形を行う。
本研究では、このような筋のメカニズムを模倣し、
ワイヤ駆動のメカニズムとして実現する。
従来のワイヤ駆動筋では、筋の起点と終点の間にワイヤを張り、
モータで巻き取ることで収縮を行わせる。しかし、
ワイヤ自体は厚みを持たないため、面状の筋をはじめとする人間の筋に近い構造はそのままでは実現できない。
また、ワイヤが直線的に張られるため、曲面形状での動作も困難である。
そこで、ワイヤ経路が、筋の始点から終点までの1本の筋繊維の経路をとりつつ、
筋肉の中でワイヤが張られていない領域も、
ワイヤの収縮に伴って、位置関係を保ちながら収縮するようなメカニズムを作成できれば、
人間の筋に近い動作を実現できると考えられる。
}%
\switchlanguage%
{%
To realize this structure, 
the requirements of the muscle structure can be described 
as the following three directions of characteristics,
as shown in \figref{figure:coord}.
\begin{itemize}
  \item x-direction: Flexibility and range of motion to realize 
  self-contraction and extension from external tension.
  \item y-direction: Strength that supports the muscle structure and allows contraction in the x-direction
  even in the presence of external load in the y-direction.
  \item z-direction: A certain degree of strength to maintain 
  the positional relationship of wire and surrounding structure  
  in the muscle, while preserving the route of wire.
\end{itemize}
}%
{%
このような構造を実現するために必要な要件は、
\figref{figure:coord}に示す3方向の圧縮・引張特性として次のようにまとめることができる。
\begin{itemize}
  \item x方向: 自身の収縮および、外部からの引張による伸びを実現できるだけの柔軟性と可動域
  \item y方向: 筋の面における外部や環境からの接触、y方向圧縮に対する一定の柔らかさと、
  接触下でも自身のx方向収縮を可能にする荷重支持能力
  \item z方向: 収縮時あるいは、他の筋により関節角度が変化した場合等の筋形状変化時において、
  筋の中におけるワイヤの経路およびワイヤを有しない部分の位置関係を保つための強度
\end{itemize}
}%
\begin{figure}[t]
  \centering
  \includegraphics[width=1.0\columnwidth]{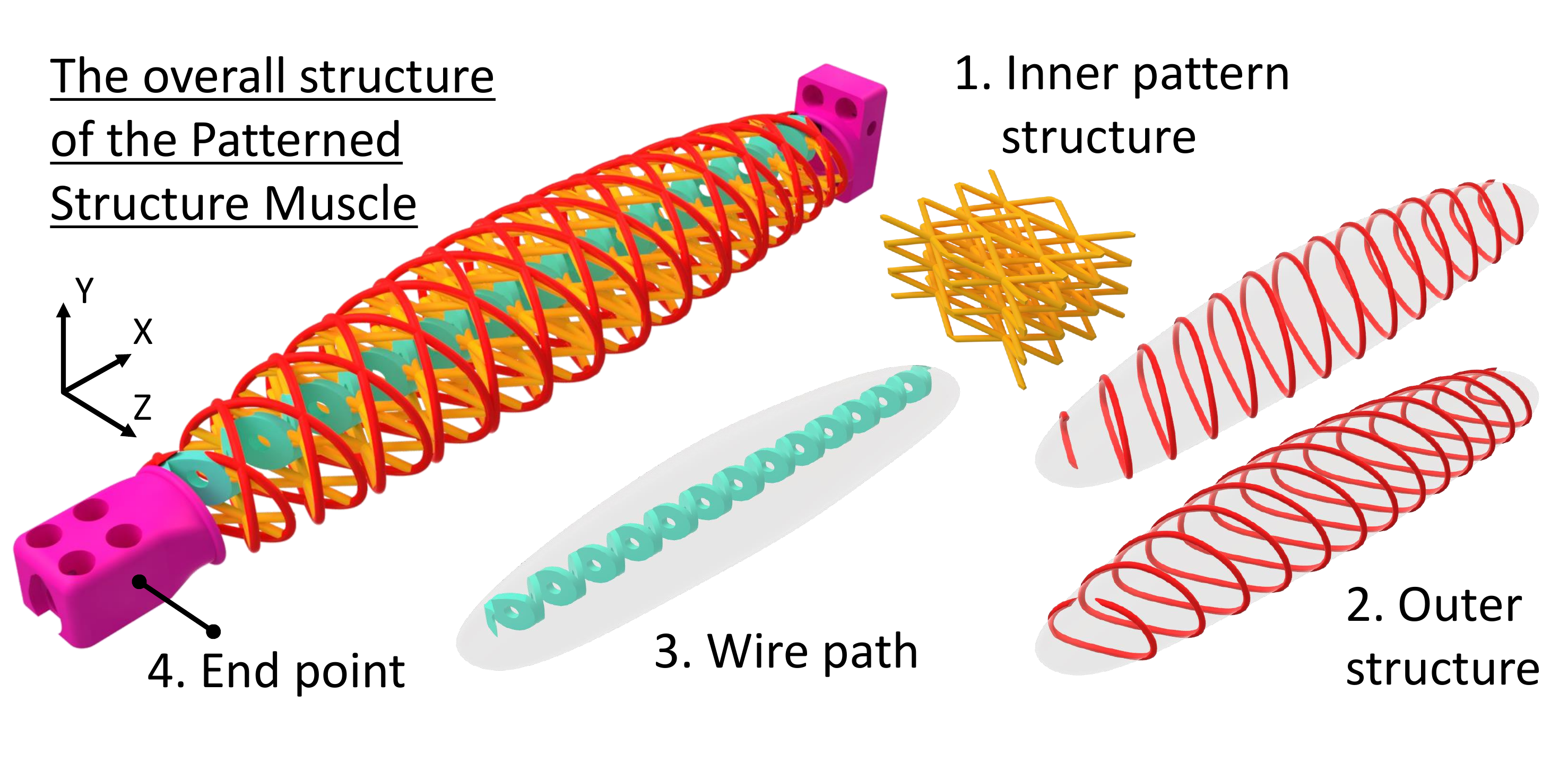}
  \vspace{-3ex}
  \caption{%
  PSM is composed of the following four structures:
  1. inner pattern structure, 2. outer structure, 3. wire path, and 4. end points.
  }
  \label{figure:structure}
\end{figure}
\begin{figure}[t]
  \centering
  \includegraphics[width=1.0\columnwidth]{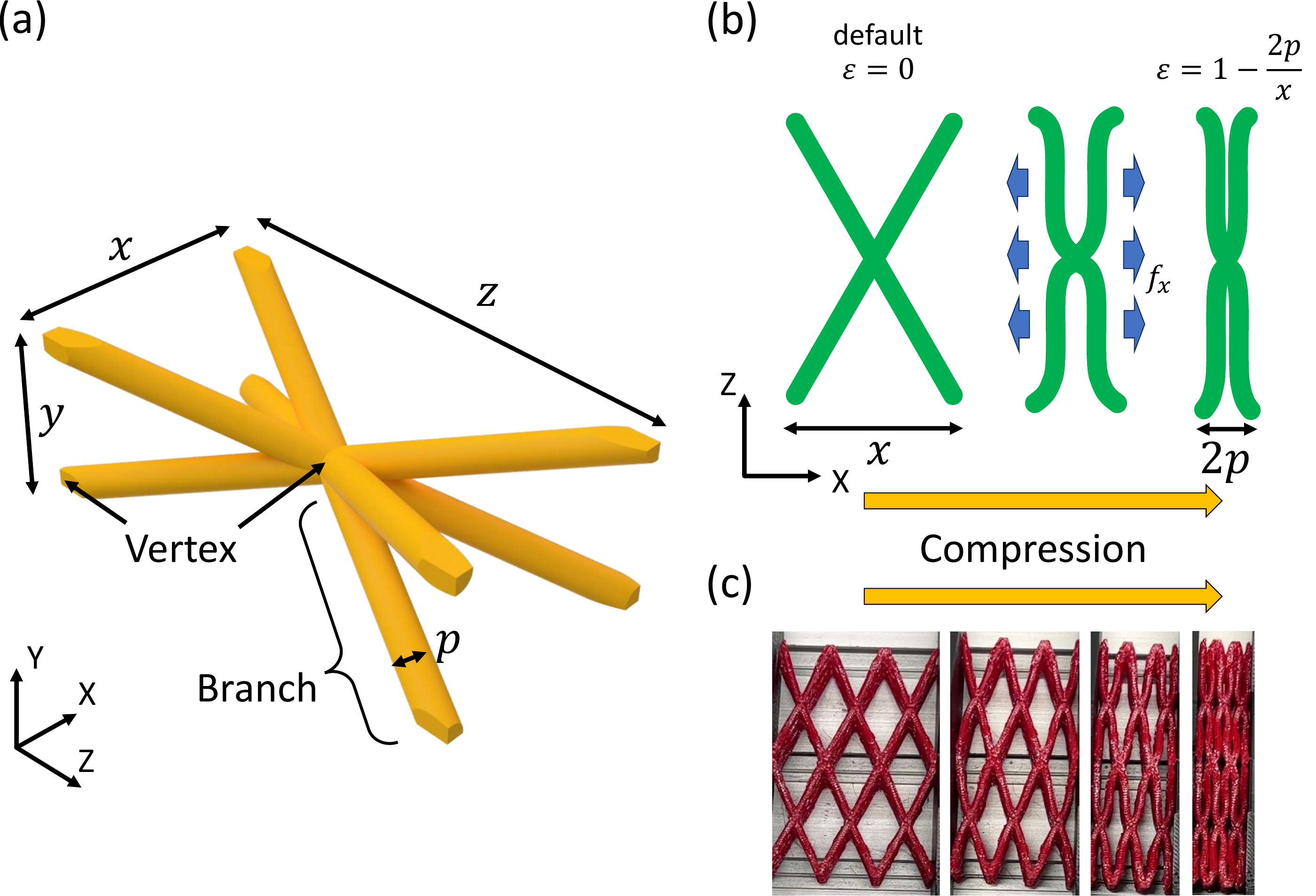}
  \vspace{-2ex}
  \caption{
  The inner pattern structure of PSM utilize the lattice structure.
  (a) illustrates the unit structure, 
  (b) illustrates the deformation during compression, and
  (c) shows actual images of the deformation.
  }
  \vspace{-1ex}
  \label{figure:infill}
\end{figure}
\subsection{Mechanism of PSM} \label{subsec:mechanism}
\switchlanguage%
{%
PSM is composed of the following four structures,
shown in \figref{figure:structure}:
1. inner pattern structure,
2. outer structure, 3. wire path, and 4. end points.
The end points are attached to the bone. 
Wire is passed through the wire path and the end points.
When the wire is pulled, the inner pattern structure, wire path, and outer structure are compressed
between the end points, resulting in contraction of the entire muscle.
The inner pattern structure, as shown in \figref{figure:infill}, 
is a critical part that has the most significant impact on the properties of PSM. 
The inner structure of PSM is formed by arranging pattern unit structures 
periodically in the x, y, and z directions. 
This pattern unit structure is composed of multiple branches.
The structure is called lattice structure, and various representative shapes exist 
\cite{seharing2020review}.
The lattice structure is well suited to additive manufacturing\cite{7790182},
and is expected to be widely used, from metal components\cite{CHEN2021100648}
to damping mechanism\cite{clough2019elastomeric}.
In the lattice structure we use, 
the length of the unit pattern in the x, y, and z directions are denoted as $x, y, z$,
as shown in \figref{figure:infill} (a).
The diameter of the branches is denoted as $p$.
These x,y,z directions correspond to the directions mentioned in \figref{figure:coord}.
The characteristics of the compression deformation are considered in two deformation phases:
}%
{%
提案する筋アクチュエータPSMの構造を説明し、\ref{subsec:concept}で述べた要件を満たしていることを示す。
PSMは、1. 内部パターン構造、2. ワイヤ経路、3. 筋表面および, 4.骨との接合部から構成され、
\figref{figure:structure}に示す。
2.ワイヤ経路および4.接合部にワイヤを通し、ワイヤを引っ張ることで、
構造1,2,3が両端の接合部に挟まれて圧縮され、
全体として収縮動作を行う。
内部パターン構造\figref{figure:infill}はPSMの性質に最も影響する主要な部位であり、
枝状構造の格子がxyzの3方向へと周期的に並べられている。
このような構造はlattice structureと呼ばれ、様々な代表形状がある\cite{seharing2020review}ほか、
Additive manufacturingとの相性がよく\cite{7790182}、
金属の構造体としての活用\cite{CHEN2021100648}から
素材の柔軟性を生かした減衰機
\cite{clough2019elastomeric}
まで、幅広く応用が期待されている。
このパターン構造における格子の3方向の長さを$x,y,z$と定め、
枝の直径を$p$とする。
なお、この$x,y,z$の方向は、\ref{subsec:concept}で述べたx,y,zの方向に対応する。
内部パターン構造の圧縮特性をモデル化し, 
パラメータを決定する規則を求める。
このとき、圧縮時の変形は、次の2つのフェーズに分けられる。
}%
\subsubsection{Bending deformation phase} \label{subsubsec:bending}
\switchlanguage
{%
Let's consider three compressions along the x, y, and z directions, respectively.
Let the nominal strain of each compression be $\varepsilon$.
Until the branches come into contact, 
deformation occurs within the range of $0 \leq \varepsilon \leq 1-{2p}/{x}, \,0 \leq \varepsilon \leq 1-{2p}/{y}, \,0 \leq \varepsilon \leq 1-{2p}/{z}$,
leading to a reduction in the angle and distance between the branches,
as shown in \figref{figure:infill} (b).
In this state, the overall compression occurs due to 
a large bending of the branches near the vertices, and
a small bending near the center of the branches.
The actual compression process is shown in \figref{figure:infill} (c).
We call this state the bending deformation phase.
In this deformation,
the magnitude relationship between $x$, $y$, and $z$ 
align with the deformation magnitudes of the branches in their respective compression directions. 
Consequently, the magnitude relationship for $x$, $y$, and $z$, 
as well as the repulsive forces per unit pattern for each direction, $f_x$, $f_y$, and $f_z$, are consistent.
Also, the larger the $p$, the larger the deformation volume,
and the larger these forces.
The average nominal stress in compression is expressed as follows:
\begin{eqnarray} 
  \sigma_{x} = \frac{f_{x}}{yz}, \,
  \sigma_{y} = \frac{f_{y}}{xz}, \,
  \sigma_{z} = \frac{f_{z}}{xy} \label{eq:fv}
\end{eqnarray}
From these expressions, we obtain the following properties:
\begin{itemize}
  \item The order of magnitudes of $x$, $y$, and $z$ corresponds to the order of magnitudes of $\sigma_x$, $\sigma_y$, and $\sigma_z$.
  \item Increasing the diameter $p$ results in larger values for $\sigma_{x}$, $\sigma_{y}$, and $\sigma_{z}$.
  \item Enlarging $x$, $y$, and $z$ individually leads to larger bending deformation regions in the corresponding x, y, and z directions.
\end{itemize}
}%
{
x方向の圧縮について考える。
x方向のひずみを$\varepsilon$とする。
まず、\figref{figure:infill}(b, c)に示すように、枝同士が接触するまで:
$0 \leq \varepsilon \leq 1-{2p}/{x}$の間、
枝の間の角度が小さくなってゆく変形をする。
一本の枝に着目すると、
枝の中心付近の緩やかな曲げ小変形と、
枝の頂点周辺の大きな曲げ変形が生じることで全体が圧縮する。
そこで、簡単なモデル化として、
初期状態における頂点間の枝部長さ$l=\sqrt{x^2+y^2+z^2}/2$, 枝の太さ$p$および、
初期状態からの枝の変形長さ差分$\Delta_x l$に応じて、
単位パターンあたりの反発力$f_{x} (l, \Delta_x l, p)$が生じると仮定する。
単位パターンにおいて、x,y,z方向それぞれ別に、
ひずみが$\varepsilon$の単方向圧縮をした場合を考える。
x,y,zの大小関係と、x,y,z方向のこれらの圧縮における枝の変形の大小関係は一致する。
よって、x,y,zの大小関係と、単位パターンあたりの反発力f_x, f_y, f_zの大小関係は一致する。
%
また、pが大きいほど、変形体積が増えて反発力$f_{x}$が大きくなる。
ここで、反発力とつり合う圧力$P_{x}$は次のように表せる。
\begin{eqnarray} 
  P_{x} = \frac{f_{x}(l, \Delta_x l, p)}{yz} \label{eq:fv}
\end{eqnarray}
y,z方向のひずみが$\varepsilon$であるようなy,z方向の圧縮に対する、
枝の変形長さ$\Delta_y l, \Delta_z l$や
圧力$P_{y}, P_{z}$も、
\eqref{eq:tt}, \eqref{eq:fv}のx,y,zを入れ替えることで求められる。
また、その場合の曲げ変形領域も, x方向と同様に、それぞれ
$0 \leq \varepsilon \leq 1-{2p}/{y}$, $0 \leq \varepsilon \leq 1-{2p}/{z}$となる。
これらより、以下の性質を得る。
\begin{itemize}
  \item $x,y,z$の大小関係は, $\Delta_x l, \Delta_y l, \Delta_z l$の大小関係と一致する。
  従って、$P_{x}, P_{y}, P_{z}$の大小関係と一致する。
  \item $p$を大きくすると、$P_{x}, P_{y}, P_{z}$は大きくなる。
  \item $x,y,z$をそれぞれ大きくすると、対応するx,y,z方向の曲げ変形領域が大きくなる。
\end{itemize}
}%
\subsubsection{Compression deformation phase} \label{subsubsec:compression}
\switchlanguage
{
When $\varepsilon > 1-{2p}/{x}$,
$\varepsilon > 1-{2p}/{y}$, and $\varepsilon > 1-{2p}/{z}$ in corresponding compression directions,
the branches come into contact, and overall compressive deformation of the material occurs.
We call this state the compression deformation phase.
In general, the pressure during compression deformation phase is larger than the pressure during bending deformation phase.

In the design of PSM,
the two phases of deformation are utilized.
By setting the parameters such that $y \leq x \leq z$,
the following properties can be obtained:
First, by reducing the value of $y$ in comparison to $x$ and $z$,
the structure can transition to the compression deformation phase with a small amount of deformation.
At the beginning of the deformation, the structure is soft, 
but after a small deformation, 
the structure becomes stiff and can support a large load.
Second, by setting the value of $x$ to be the second largest,
the contraction distance of x direction in the bending deformation phase is made larger than that in the y direction,
and the repulsive force of x direction, $f_x$, during the bending deformation phase is 
not as large as that in the z direction.
This means that the structure is flexible in the x-direction, enabling substantial contraction over long distances.
Finally, by setting the value of $z$ to be the largest,
the pressure during bending deformation in the z direction is made the largest.
This ensures a certain degree of structural support strength to maintain the wire path and overall shape.
These properties satisfy the requirements of x, y, and z directions mentioned in \ref{subsec:concept}.
}
{%
次に、$\varepsilon > 1-{2p}/{x}$となり、枝同士が接触した状態では、
素材全体の圧縮変形が生じる。
この圧縮時の圧力は、一般に、曲げ変形時の圧力$P_x$よりも大きい. 

以上の、曲げ変形・圧縮変形という2つのフェーズに分けた変形を活用する。
PSMにおいては、$y \leq x \leq z$となるようなパラメータを設定することで、
次のような性質を得ることができる。
まず、yの値は最も小さくすることで、少しの変形で圧縮変形のフェーズに移行することができ、
変形の最初は柔らかいものの、一定の変形を超えると急激に弾力が増し、大きな荷重を支持する。
次に、xの値は2番目に大きくすることで、曲げ変形の収縮距離をy方向と比べて大きくすることができ、
かつ、曲げ変形時の反発力はz方向ほど大きくならないために柔軟に収縮する。
最後に、zの値は最も大きくすることで、曲げ変形時の反発力を最も大きくし、
ワイヤの経路および全体の形状を保てる程度の、一定の構造支持強度を確保する。
これらは\ref{subsec:concept}で述べたx,y,z方向の要件を満たす。
}%
\switchlanguage%
{%
While utilizing the characteristics of the inner pattern structure described above,
the wire path, outer pattern as a muscle surface and endpoints for connection with the bone are realized.
The wire path and outer pattern are arranged by taking the y-directional projection of the inner pattern 
based on the shapes of the tubes for threading wires and the muscle surface, as illustrated in \figref{figure:structure} (3, 2).
This allows these parts to function without interfering with the x-directional deformation,
y-directional load support, and z-directional internal structural support.
Also, the end points (\figref{figure:structure} (4)) are composed of simple dense structure that is suitable for attachment.
In summary, PSM utilizes the three directions of anisotropic characteristics, 
and the combination of sparse lattice structure and dense parts, 
to achieve various properties with a single material.
}%
{

上で述べた内部パターン構造の特性を生かしながら、
ワイヤの配置、筋表面の滑らかさの確保、骨との接合を実現する。
ワイヤ経路および筋表面は、\figref{figure:structure}に示すように、
ワイヤを通す管および筋表面の形状を元に、
内部パターンのy方向の射影をとって配置される。
これによって、これらの部位が、x方向の変形、
y方向の荷重支持、z方向の構造支持強度、それぞれの性能をできるだけ損わず動作する。
また、骨との接合部は
密な単純構造であるため、同じ材料でありながら硬くて接合に適する。

このように、PSMにおいては、3方向の可動域・弾性の異方性の利用および、
疎な部位と密な部位を組み合わせることで、
1種の素材で様々な特性を実現している。
}%

\subsection{Process of Making PSM}
\begin{figure}[t]
  \centering
  \includegraphics[width=1.0\columnwidth]{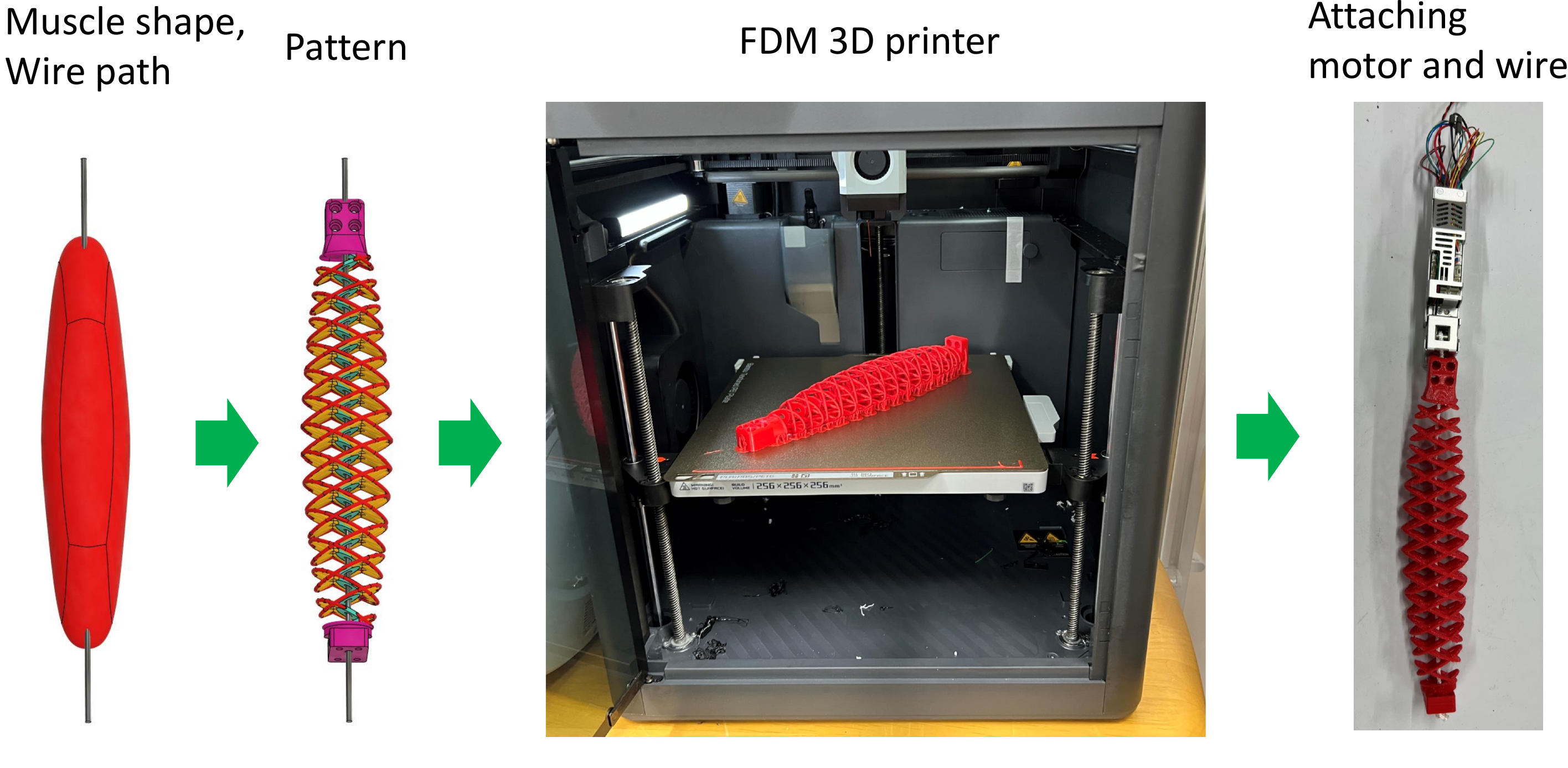}
  \vspace{-2ex}
  \caption{%
  The making process of PSM consists of four steps:
  1. modeling muscle shapes and wire path,
  2. converting them into pattern structure,
  3. FDM 3D printing, and 4. attaching wires and motors.
  }
  \vspace{-1ex}
  \label{figure:procedure}
\end{figure}

\switchlanguage%
{
The process of making PSM is shown in \figref{figure:procedure}.
This process consists of modeling, 3D printing, and attaching wires and motors.
In the modeling process, the overall shape of the muscle and the route of the wire are determined firstly,
and then they are converted into a pattern structure.
We generate pattern structure from the shape, route and structural parameters $x,y,z,p$ by a script.
Then, PSM is created by FDM 3D printing using Thermoplastic Polyurethane (TPU) as the material.
Finally, wires and motors are attached to complete the assembly.

Using FDM 3D printing,
it is possible to create various shapes.
Also, it is relatively inexpensive, readily available, and easy to use,
and does not require special skills or post-processing.
}%
{%
PSMの製造方法を\figref{figure:procedure}に示す。
この製造方法は、モデリング、積層3Dプリント、ワイヤおよびモータの取り付け、という単純なプロセスからなる。
モデリングにおいては、筋の全体形状およびワイヤの経路を決定したのち、
それらをパターン構造に変換する。
$x,y,z,p$のパラメータを定めれば、これらの形状・経路から、
PSMのパターン構造は機械的に容易に生成可能であり、
我々はスクリプトによりこれを生成している。
骨との接合部のみ、手動のモデリングを行った。
TPUの積層3Dプリントにより、このパターン構造を実際の物体として作成し、
最後ワイヤ・モータを取り付けて完成する。
この製造方法は、
積層3Dプリント造形のため、
様々な形状に対応可能であるうえに、
比較的安価で入手性が高く、かつ使用が容易であり、
特別な技術や後処理などの手間が必要ない。
}%

\subsection{Various Shapes of PSM and Upper Arm Using PSM}
\begin{figure}[t]
  \centering
  \includegraphics[width=1.0\columnwidth]{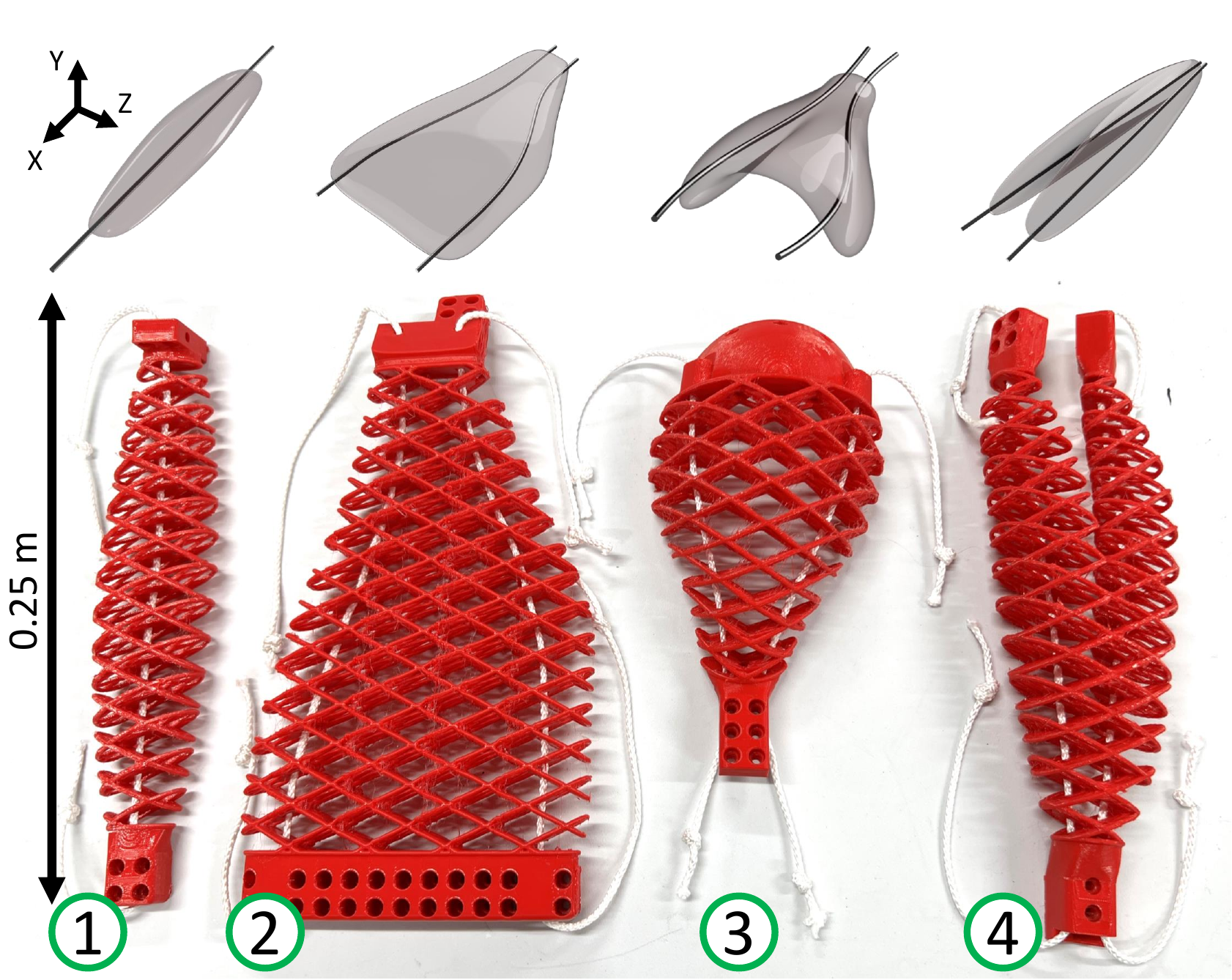}
  \vspace{-2ex}
  \caption{%
  The four shapes of PSM: Muscle 1 is a simple single-wire muscle, 
  muscle 2 is a wide sheet-like muscle. 
  muscle 3 is a muscle covering a joint, 
  and muscle 4 is a branched muscle.
  muscle 2,3,4 are 2-DOF muscles.
  }
  \label{figure:shape}
\end{figure}

\begin{figure}[t]
  \centering
  \includegraphics[width=1.0\columnwidth]{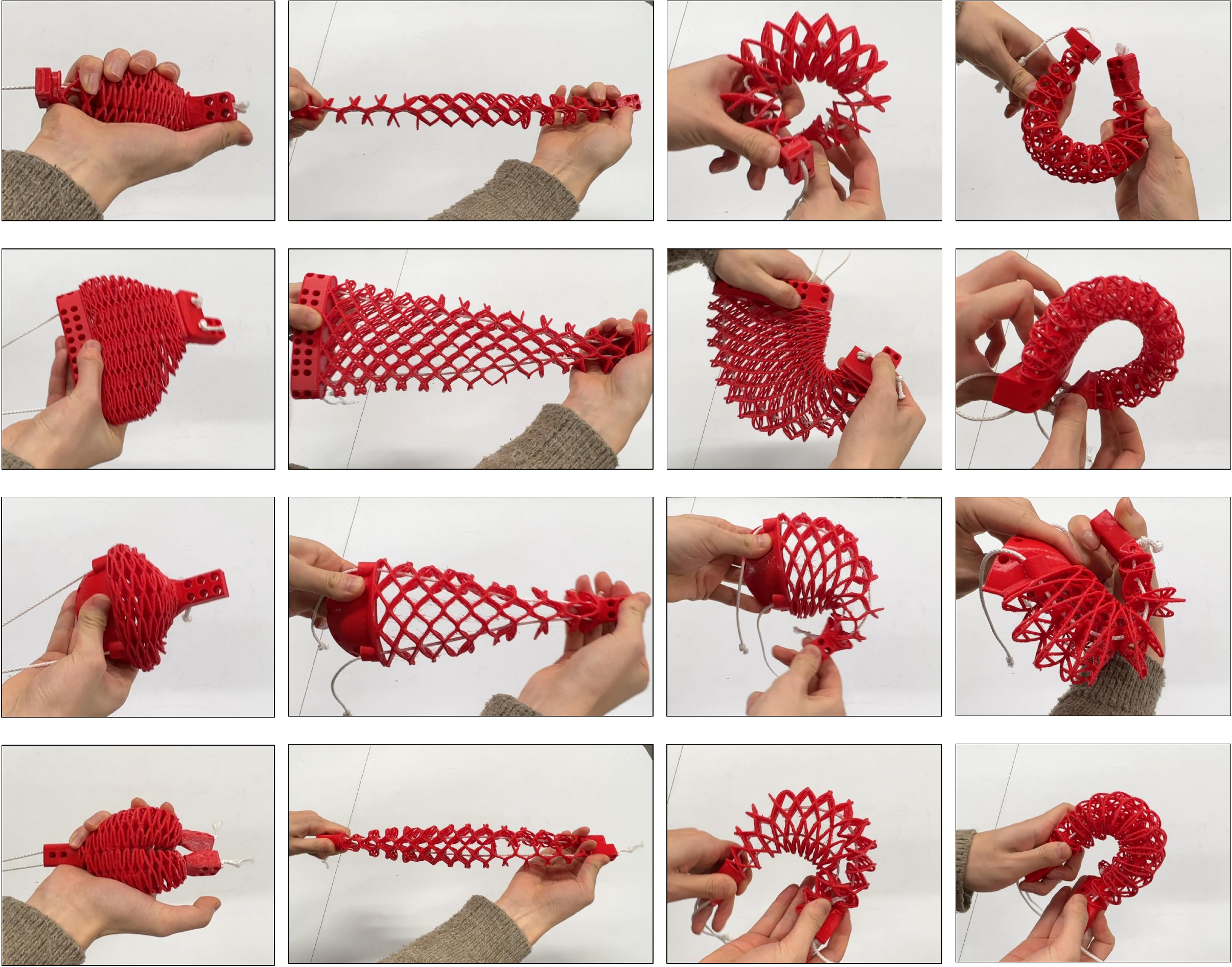}
  \caption{%
  The deformation of muscles 1, 2, 3, and 4 are illustrated from top to bottom. 
  In the leftmost column, the deformation caused by pulling the wire are shown.
  On the right side, the stretching deformation in the x-axis, and the bending deformation in the y and z axes are shown.
  }
  \vspace{-2ex}
  \label{figure:deformation}
\end{figure}

\begin{figure}[t]
  \centering
  \includegraphics[width=1.0\columnwidth]{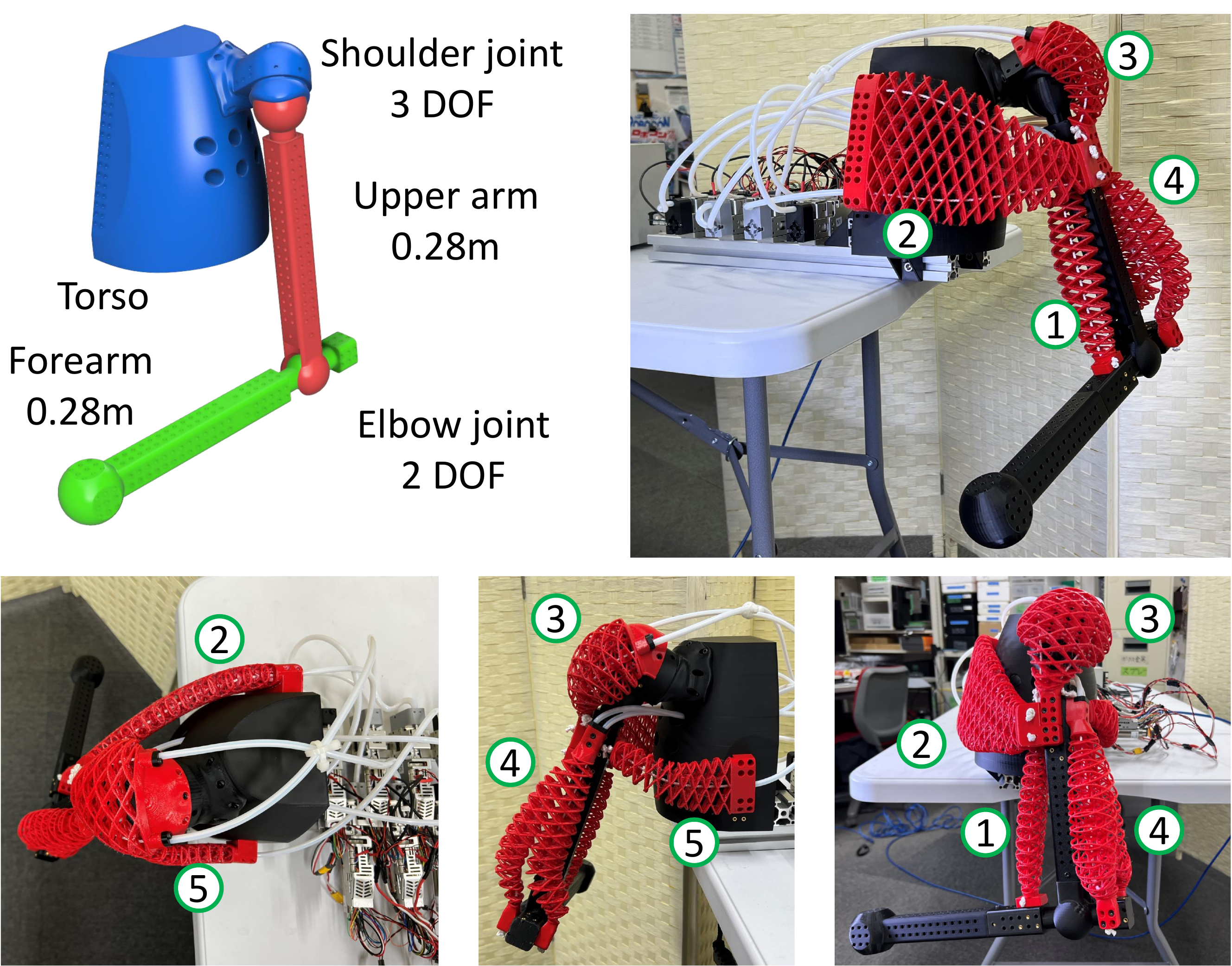}
  \vspace{-2ex}
  \caption{
  The upper arm structure combining PSMs.
  Muscle 1 is positioned for the biceps brachii, 
  muscle 2 for the pectoralis major,
  muscle 3 for the deltoid,
  and muscle 4 for the triceps brachii.
  An additional muscle (muscle 5) is added as the back muscle.
  Note that this design was created to demonstrate the feasibility of using PSM as muscles,
  and is not intended to strictly mimic the human musculoskeletal structure.
  }
  \label{figure:arm}
\end{figure}

\switchlanguage%
{%
Utilizing the basic characteristics of PSM described so far, 
we created four types of PSM. 
As shown in \figref{figure:shape},
these are, muscle 1: simple single muscle, 
muscle 2: wide sheet-like muscle, 
muscle 3: muscle that covers a joint, 
and muscle 4: branched muscle.
Except for the simple single muscle, all of them are 2-DOF muscles. 
The muscle that covers a joint has a round shape to cover the interior,
and the branched muscle is created to have two branches that contract separately.
The pattern parameters were standardized to (x, y, z, p) = (15, 10, 30, 2) (mm). 
Specifically, the specifications of a simple single muscle is presented in \tabref{table:spec}.
These muscles flexibly deform in response to external forces. 
In particular, photographs illustrating the deformation of PSMs by hand
are presented in Figure \ref{figure:deformation}.
While flexibly deforming in response to external forces, they also protect the wire and its path.

Furthermore, we created an upper arm structure, as shown in \figref{figure:arm}, 
by combining these muscles. 
It consists of bones and joints printed with PLA, and PSMs made from TPU.
The motor-wire units from previous research\cite{Musashi} are used,
and the wires are passed from the motors to the muscles through PTFE tubes.
The joint structure includes a 3-DOF ball joint at the shoulder,
and a 2-DOF joint at the elbow that enables flexion, extension, as well as internal and external rotation.
The muscle 1,2,3,4 described above are placed at the positions of the 
biceps brachii, pectoralis major, deltoid, and triceps brachii.
In addition, a single muscle (muscle 5) is added as the back muscle.
This structure was created to demonstrate the potential of PSM,
and is not intended to strictly mimic the human musculoskeletal structure.
}%
{%
これまで説明してきたPSMの基本的な構造を用いて、4種に分類されるPSMを作成した。
\figref{figure:shape}に示すように、
単純な1筋、広い面状の筋、関節を覆う筋、分岐した筋の4種である。
単純な1筋以外はいずれも多自由度筋である。
また、関節を覆う筋は、丸い形状で関節を覆うように作成されており、
分岐した筋は、2つの筋が分岐し、それぞれが別々に収縮するように作成されている。
パターンパラメータは、
(x,y,z,p) = (15, 10, 30, 2)\,[mm]で統一した。
特に、単純な1筋について、性能を\tabref{table:spec}に示す。
これらの筋は、外力に対して柔軟に変形する。
特に、手でワイヤを引っ張る、PSMを引っ張る・曲げた状態の写真を\figref{figure:deformation}に示す。

さらに、これらの筋を組み合わせた上腕構造\figref{figure:arm}を作成した。
PLAのプリントで作成された骨および関節とTPUのPSMから構成される。
モータは先行研究\cite{Musashi}のワイヤユニットを使い、
PTFEチューブでワイヤをモータから筋へと通している。
肩の3自由度球関節および肘の2自由度関節を有する関節構造に、
上記の筋を上腕二頭筋および背筋、大胸筋、三角筋、上腕三頭筋として配置した。
なお、PSMの利用可能性を示すために作られた構造であり、
厳密に人体の筋骨格構造を模倣したものではない。
}%
\begin{table}[t]
 \begin{center}
  \caption{Specifications of muscle 1}
  \begin{tabular}{cc} \toprule
  Parameter & Value \\ \midrule
  Pattern size parameter (x,y,z) & (15, 10, 30) [mm] \\
  Pattern branch diameter  & 2 mm \\
  Muscle size (X,Y,Z) & (250, 30, 50) [mm] \\
  Length after contraction in the x-axis & 115 mm \\
  Length after stretching in the x-axis & 380 mm \\
  Range of twisting in the x-axis & over $\pm$360 deg \\
  Bending in the y and z axes & over $\pm$360 deg \\
  Mass of PSM & 38 g \\
  Material of PSM & TPU 95A \\
  Maximum tension of the motor-wire unit & 300 [N] \\
  Maximum speed of the motor-wire unit & 250 [mm/s] \\
  \bottomrule
  \end{tabular}
  \label{table:spec}
 \end{center}
\vspace{-4ex}
\end{table}

\section{Experiments}\label{sec:experiments}
\subsection{Measurement of PSM Properties With Varied Parameters} \label{subsec:measurement}
\begin{figure}[t]
  \centering
  \includegraphics[width=1.0\columnwidth]{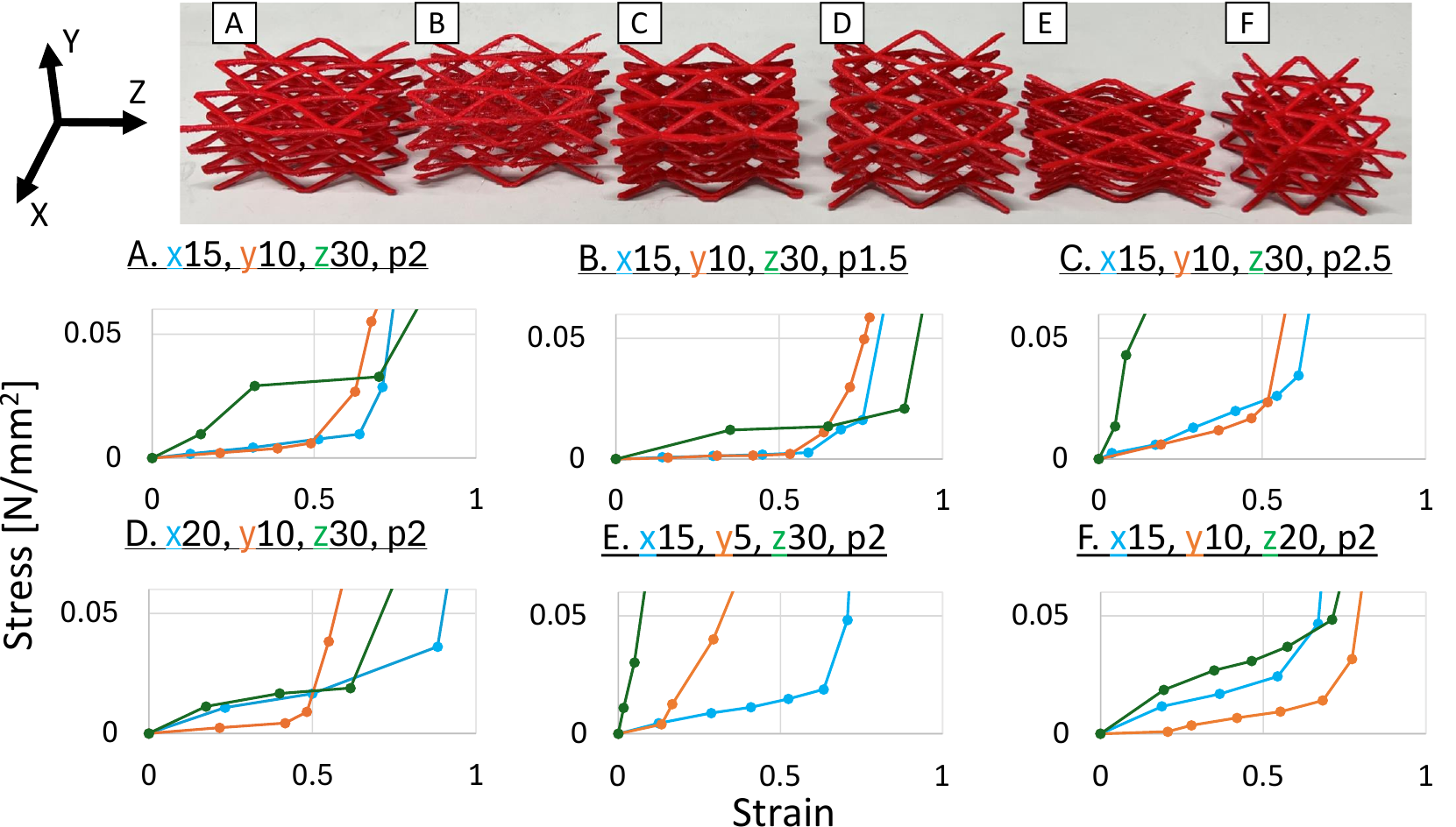}
  \caption{%
  Compression measurements on internal structures. 
  Displaying nominal stress-strain curves for six inner structures with different pattern parameters.
  }
  \vspace{-2ex}
  \label{figure:push}
\end{figure}

\begin{figure}[t]
  \centering
  \includegraphics[width=1.0\columnwidth]{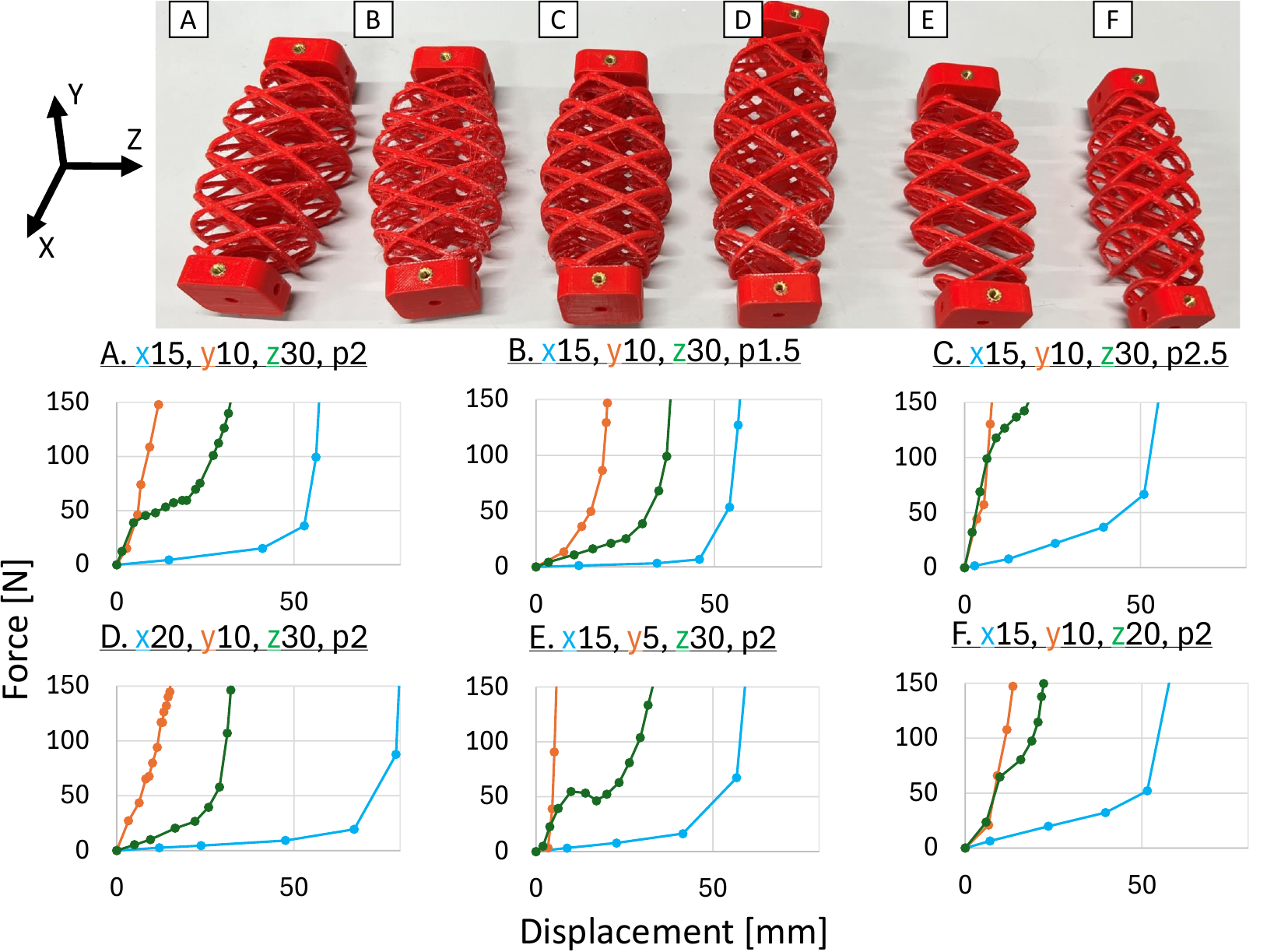}
  \caption{%
  Compression measurements on PSMs.
  Displaying force-displacement curves for six muscles with different pattern parameters.
  }
  \vspace{-2ex}
  \label{figure:push_muscle}
\end{figure}

\switchlanguage
{%
First, we measure the anisotropic characteristics of PSM.
We created six inner structures and PSMs with different pattern parameters $(x,y,z,p)$,
to measure their compression characteristics.
Condition A has the reference parameters (15, 10, 30, 2) (mm), 
and B, C have thickness p as 1.5 and 2.5, respectively. 
D, E, F have altered x, y, z to 20, 5, 20 from A's parameters, respectively.
For measurements, inner structures with pattern numbers of 3, 3, 2 for x, y, z were used.
The characteristics of PSM were measured 
using muscles with dimensions of approximately 75 x 30 x 45 (mm). 
In the case of D, E, F, the size of the entire muscle shapes was multiplied 
by the ratio of the pattern length,
and the pattern numbers and arrangement were the same as A.

In \figref{figure:push}, the nominal pressure-strain curves of the internal structure are shown. 
The color of each line in the graph corresponds to the color of the x,y,z in the graph title.
First, especially for the x and y directions, 
it can be observed that the pressure increases rapidly when the strain reaches around 0.6, 
indicating a transition from the bending deformation phase to the compression deformation phase.
Generally, the pressure during bending deformation is large depending on the size of x, y, and z.
It means that adjusting these size relationships makes it possible to 
control the anisotropic characteristics.

In \figref{figure:push_muscle}, the force-displacement curves of PSMs are shown.
In the x and z directions, bending deformation occurs over a wide area. 
The x-direction is soft and suitable for contraction, 
while the z-direction is relatively rigid and suitable for maintaining shape.
In the y-direction, there is a transition to compression deformation phase with slight deformation, 
making it the strongest. 
This is due to not only the size of y, but also the presence of internal wire paths, 
which reduce the bending deformation region of the y-direction.
In summary, it is shown that PSM has anisotropic characteristics utilizing the pattern structure,
and also that these characteristics can be adjusted by changing the pattern parameters.
}
{
まず、\ref{subsec:mechanism}で述べたような異方的な特性を、PSMが有することを測定により示す。
パターン構造のパラメータ$(x,y,z,p)$が異なる6種の内部構造およびPSMを作成し、
圧縮特性を測定した:
Aが基準パラメータ(15, 10, 30, 2) [mm]であり、
B,Cはそれぞれ太さpが1.5および2.5と太い。
D,E,Fはそれぞれx,y,zを20, 5, 20へと変えたものである。
内部構造特性は、x,y,zのパターン数が3,3,2の構造を作成して,
最大40mmの圧縮変形により計測し、圧力とひずみに変換した。
PSMの特性は、大きさがおよそ75x30x45 [mm]である筋をAのパラメータで作成し、
B,Cは太さのみを変えた。D,E,Fはそれぞれパターン長さに応じて、
対応する各方向の筋全体形状の大きさに倍率をかけており、
パターン数およびパターン配置はAと同じである。

\figref{figure:push}に、内部構造の圧力-ひずみ曲線を示す。
各グラフの線の色が、タイトルのxyzの色に対応する。
まず、特にx,y方向について、ひずみが0.6前後になると圧力が急激に増加し、
曲げ変形から圧縮変形に移行していることがわかる。
また、z方向の曲げ変形領域においてはx,yよりも大きな力が生じていることがわかる。
基本的に、x,y,zの大きさに応じて、曲げ変形時の圧力の勾配が大きく、
x,y,zの大小関係を変えることで異方性を調整可能であるとわかる。
一方、Eではx方向の曲げ変形時の圧力が最も小さいが、
Eにおいてはyが非常に小さく、
積層方向がyであるためにy方向の枝同士が付着してしまい、
強度がより高くなっているためであると考えられる。

\figref{figure:push_muscle}に、PSMの変位-力曲線を示す。
曲げ変形、圧縮変形という2つのフェーズに分かれた変形により、
小さな変形では柔らかいが、大きな変形になると急激に弾力が増していることがわかる。
なお、ワイヤ経路(\figref{figure:structure}(3))が内部に存在することの影響により、
y方向の曲げ変形領域は小さくなり、
少しのy方向変形で圧縮変形に移行しているほか、強度も高くなっている。
これらにより、PSMが、内部のパターン構造を活用した特性を有していること、
パラメータの変化によりその特性を調整可能であることが示された。
}%
\subsection{Contraction Motion of PSM} \label{subsec:psm_move}
\begin{figure}[t]
  \centering
  \includegraphics[width=1.0\columnwidth]{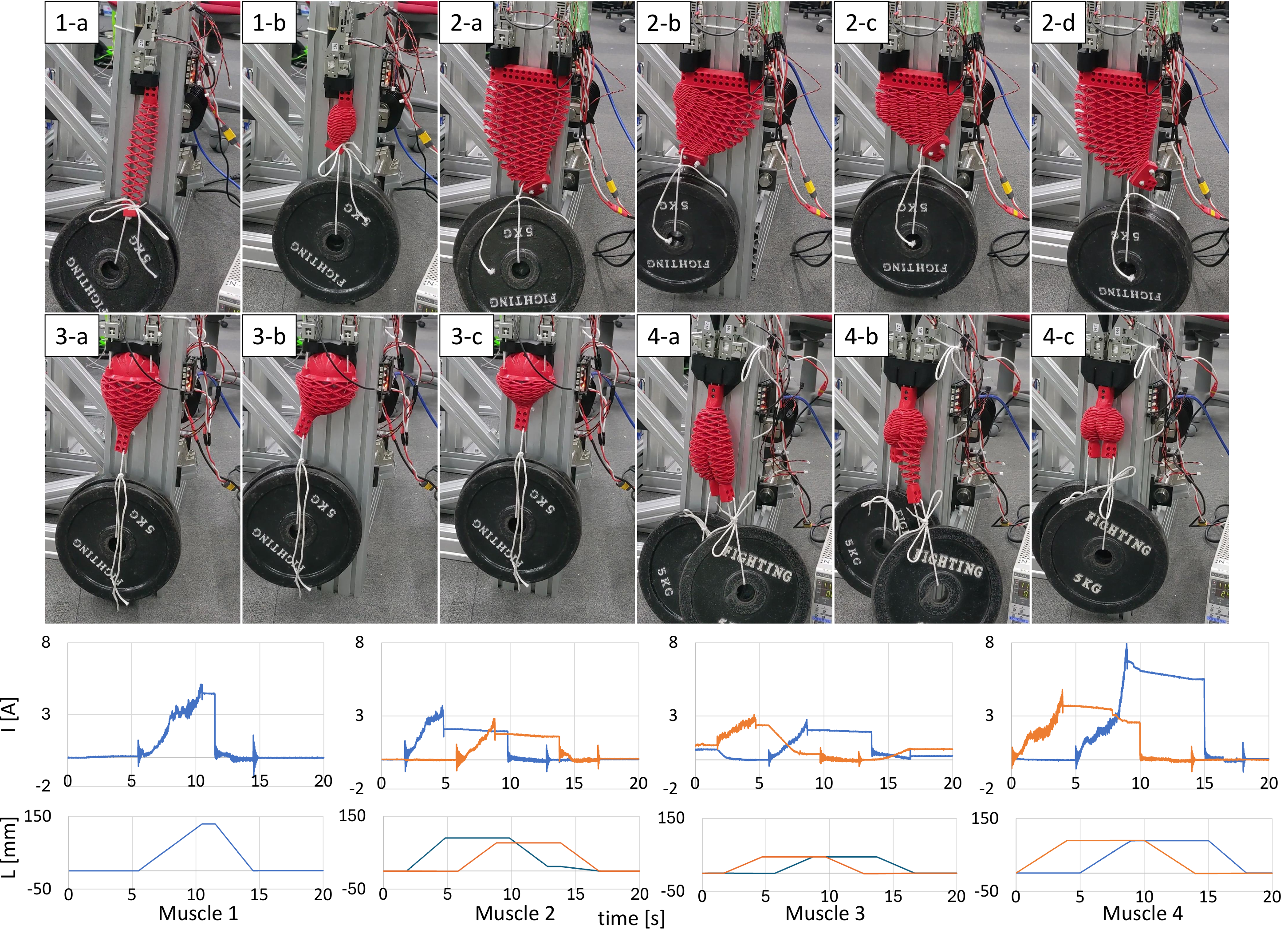}
  \vspace{-2ex}
  \caption{
  With each shape of PSM shown in \figref{figure:shape},
  a total of 10kg weight was lifted. 
  In the 2-DOF muscles, two wires were operated independently.
  }
  \label{figure:weight}
\end{figure}

\begin{figure}[t]
  \centering
  \includegraphics[width=1.0\columnwidth]{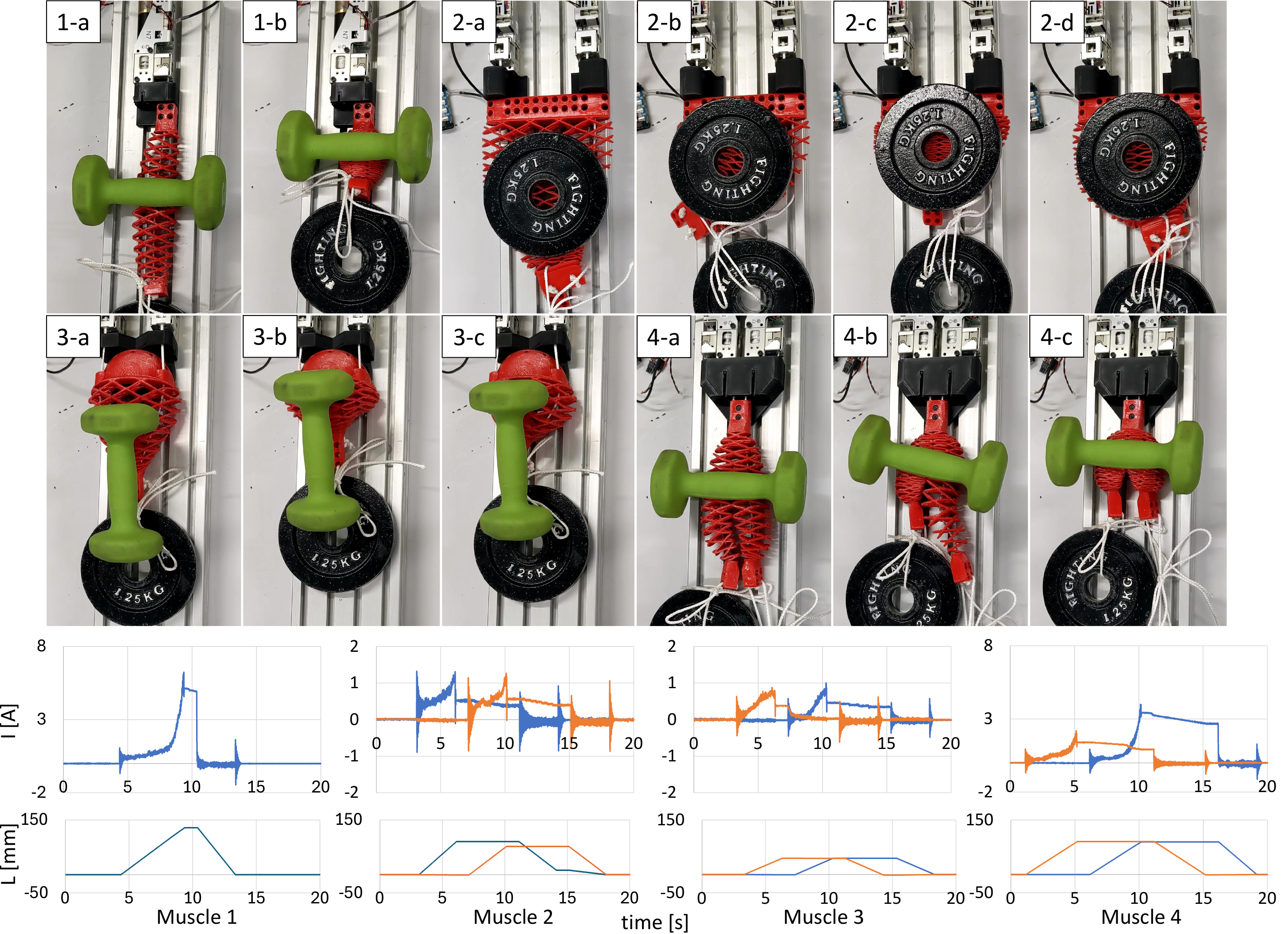}
  \vspace{-2ex}
  \caption{%
  We performed the same contraction motion as in \figref{figure:weight}, 
  with the muscle placed horizontally and the weight leaning against it, 
  confirming its operation under contact. 
  The green weight is 1.38kg, and the black weight is 1.25kg.
  Each of the PSMs pulls one or two 1.25kg weights.
  }
  \vspace{-2ex}
  \label{figure:weight_on}
\end{figure}

\switchlanguage
{%
Next, the performance of the four types of PSM created (\figref{figure:shape}) 
are tested in two different environments. 
To confirm its ability to exert large contraction force, 
a total weight of 10 kg was lifted using the PSM,
as shown in \figref{figure:weight}.
Also, to demonstrate its movements under contact, 
weights of 1.25 kg or 1.38 kg were placed or leaned on the horizontally positioned PSM,
and the contraction was performed, as shown in \figref{figure:weight_on}.
At the top of the each figure, the PSMs' contraction are shown.
The muscle number and time are shown as 1-a, 1-b, etc.
At the bottom, the current and wire length of each muscle are shown.
In muscles with multiple degrees of freedom, such as Muscle 2, 3, 4,
the contraction is performed by contracting one side first, and then both sides,
and finally the opposite side.
The plots for the 2-DOF muscles are overlaid in the graph.
In the initial experimental setup, 
it was confirmed that the PSM is capable of lifting heavy objects. 
Furthermore, in the second experimental environment, 
it was verified that the PSM can operate under the load from the environment.
In muscles 2, 3, and 4, 
2-DOF movements such as contraction on only one side, 
contraction on both sides simultaneously, 
were performed, 
utilizing the interaction of the two wires. 
In these cases, the structure around the wire paths and the wire paths themselves 
undergo deformation while maintaining their internal positional relationships,
as described in \figref{figure:concept}.
These results show that PSM can utilize the advantages of wire-driven actuation,
such as the ability to exert large forces and control the length of the wire accurately,
depending on the motor.
Additionally, it demonstrates that even in the presence of environmental contact, 
the deformation of wire paths remain stable, 
overcoming the weaknesses of wire-driven actuation.
}
{%
作成した4種のPSM(\figref{figure:shape})の動作を2つの環境で確認する。
重量物持ち上げ可能であることの確認として、\figref{figure:weight}に示すように、
合計10kgの重りをPSMで持ち上げた。
また、\figref{figure:weight_on}に示すように、
横に置いたPSMに1.25kgまたは1.38kgの重りを載せる、立てかけるなどの方法で
面に荷重をかけながら、重りを引っ張る様子を示す。
図の上部には、\figref{figure:shape}で示した4種のPSMが、
合計10kgのおもりを吊るした状態での収縮運動を行っている写真を示し、
筋番号-時間経過を1-a, 1-b,...などと示した。
図の下部には、それらの筋の電流およびワイヤ長さ$I, L$の時間変化を示し、
ワイヤを2本持つ筋は、それらの変化を重ねて描画した。
最初の実験環境では、PSMが重量物を持ち上げることができることが、
また、2つ目の実験環境では、PSMが環境からの荷重下での動作を行うことができることが確認された。
また、筋2,3,4のような多自由度筋においては、その自由度を生かし、
まず片側のみ収縮させ、次にもう片方を収縮させている。
その場合も、ワイヤ経路のまわりの構造が、
\figref{figure:concept}で述べたように位置関係を保ちながら変形している。
これらにより、PSMが、モータ依存で大きな力を発揮できるワイヤ駆動の強みを生かしていること、
接触があってもワイヤ経路が安定して変化し、ワイヤ駆動の弱点を克服できている
ことが示された。
}%

\subsection{Motion of The Arm Constructed with PSM} \label{subsec:arm_move}
\begin{figure}[t]
  \centering
  \includegraphics[width=1.0\columnwidth]{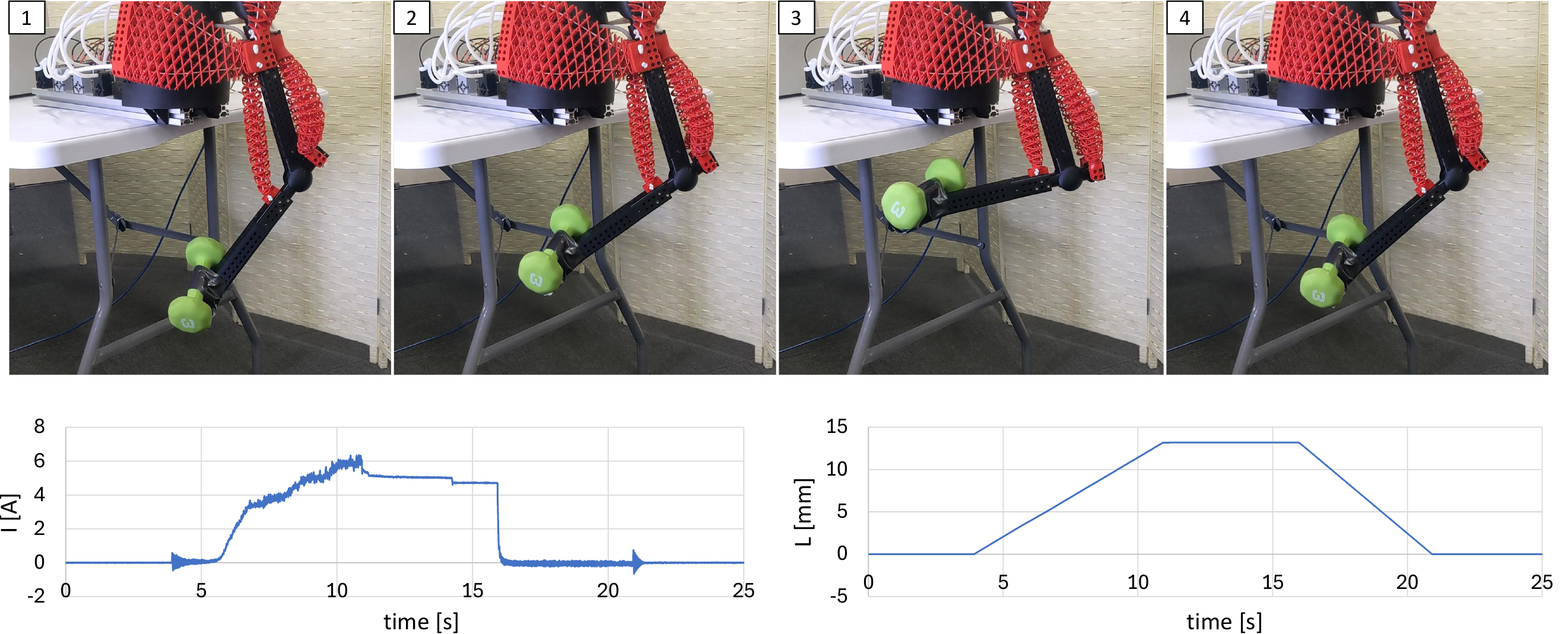}
  \vspace{-2ex}
  \caption{%
  We conducted a motion experiment lifting a 1.38kg weight attached to the hand 
  using muscle 1 at the position of the biceps brachii in the upper arm.
  We only plot the current and length of muscle 1 in this experimental setup,
  because other muscles' current and length are almost constant.
  }
  \vspace{-2ex}
  \label{figure:arm_weight}
\end{figure}

\begin{figure}[t]
  \centering
  \includegraphics[width=1.0\columnwidth]{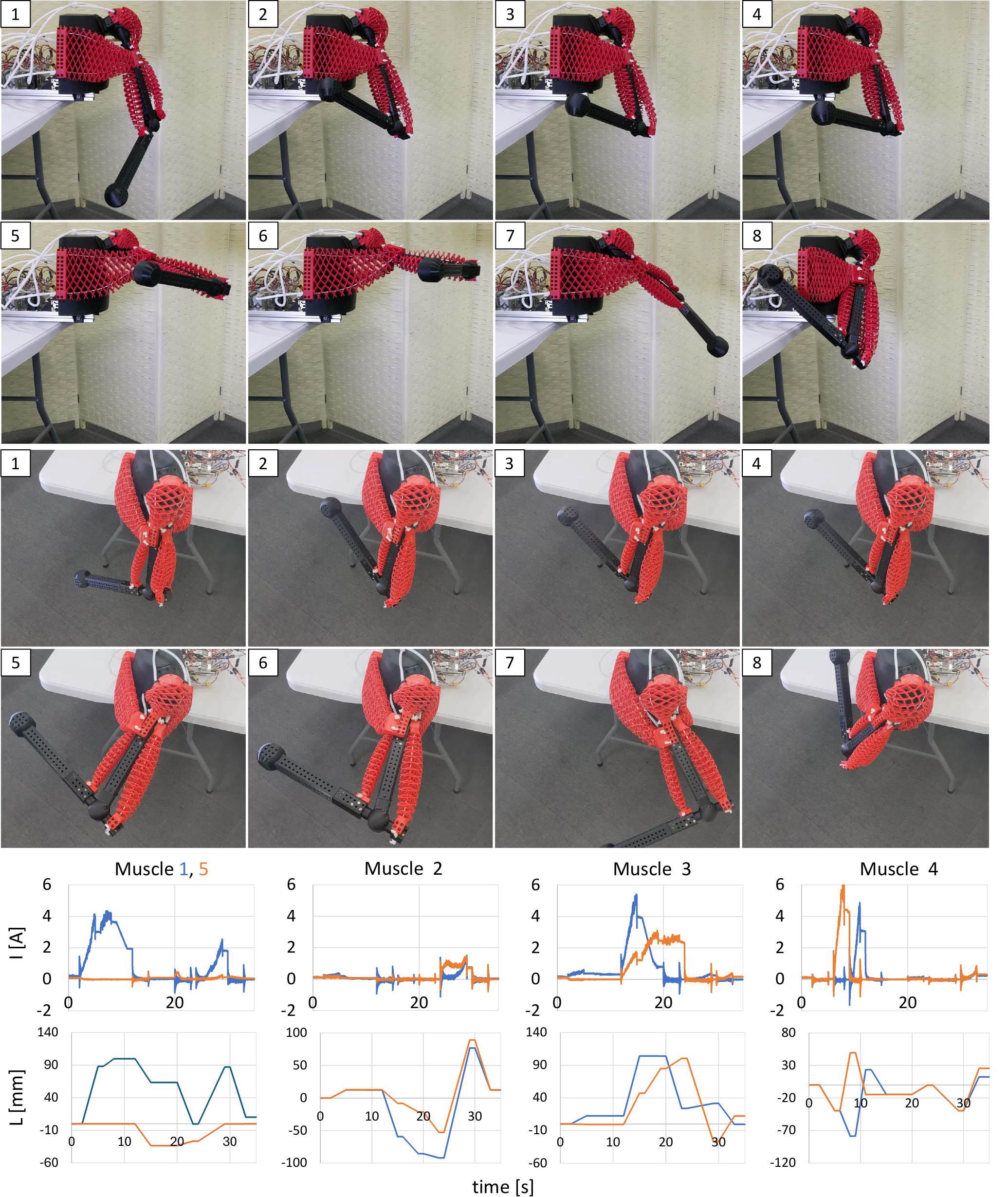}
  \vspace{-2ex}
  \caption{
  A series of movements are performed using the upper arm structure.
  It consists of 8 postures, and the motion is performed for about 30 seconds.
  The arm utilizes muscles such as covering joints and branched muscles,
  and the wide range of motion is achieved.
  The muscles interact with each other, undergoing bending and tensile deformation
  by the contraction of other muscles,
  yet they still function as muscles under these conditions.
  }
  \label{figure:arm_merge}
\end{figure}

\begin{figure}[t]
  \centering
  \includegraphics[width=1.0\columnwidth]{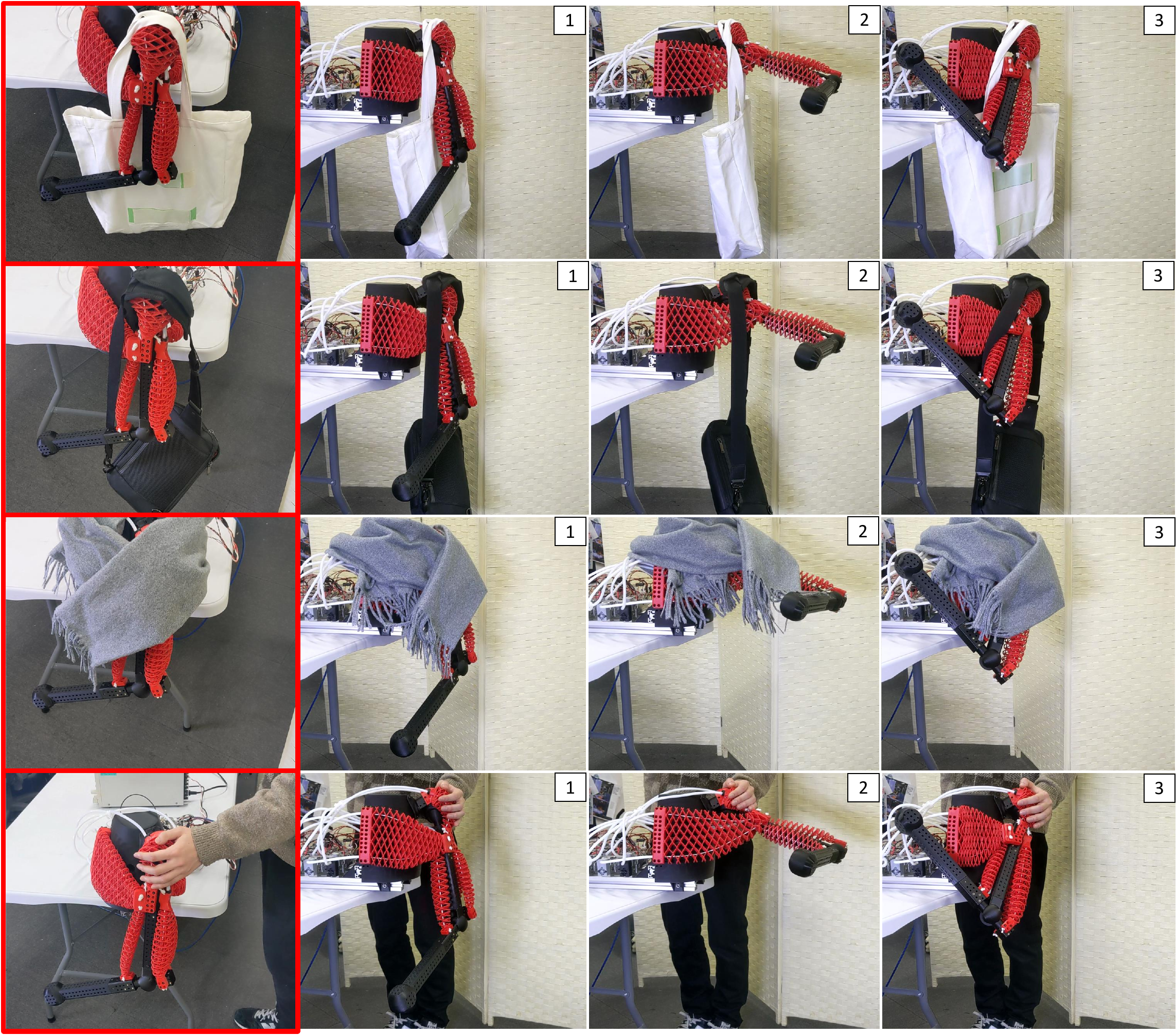}
  \caption{
  Upper arm motion in situations where muscles come into direct contact with the environment. 
  Even in scenarios involving contact with bags, fabric, or hands, 
  the muscles operate safely while withstanding external contact loads.
  }
  \label{figure:arm_env}
\end{figure}

\switchlanguage
{%
In this experiment, we demonstrate the operation of the upper arm structure shown in \figref{figure:arm}.
Firstly, we perform the motion of lifting a weight using the muscle 1,
located at the position of the biceps brachii, as shown in \figref{figure:arm_weight}.
We send position commands to the motor and make the elbow flex 
with a 1.38kg weight attached to the hand.
The flexion is carried out over 7 seconds, making the forearm almost horizontal.
The arm keeps the position for 5 seconds, and then extends over 5 seconds.
When calculated from the moment arm of the muscle 1, a load of about 200N 
is applied to the muscle. 
This is within the capacity of the motor unit used, and the lifting is successful.
As shown in \ref{subsec:psm_move}, the tensile load of PSM is supported by the motor and wire.
Therefore, PSM can exert a large tensile force according to the performance of the motor.
This capability is also effective in applications for robots,
such as the motion of lifting a weight in the upper arm.
}
{%
\figref{figure:arm}で示した上腕構造について動作を行わせる。
まず、肘の上腕筋を収縮させておもりを持ち上げる動作を\figref{figure:arm_weight}に示す。
モータに位置司令を送り、1.38kgの重りを手に付けた状態での肘の屈曲を行った。
初期角度から45度の屈曲を7秒で行い、前腕がほぼ水平になった状態で5秒間保持し、
その後5秒で腕を伸展させた。
上腕筋のモーメントアームから計算すると、筋には200Nの荷重がかかっているが、
これは使用したモータユニットの定格内であり、持ち上げに成功した。
このように、PSMの引張方向の荷重はモータおよびワイヤが支持するために、
モータの性能に応じて大きな引張力を発揮できることが示された。
}
\switchlanguage
{%

Next, we perform a motion of the entire upper arm for about 30 seconds, 
as shown in \figref{figure:arm_merge}.
The top of the figure shows the motion,
and the bottom shows the current and wire length of each muscle.
The plots for Muscle 1 and Muscle 5 as described in \figref{figure:arm} were overlaid.
The format of the plots for the other muscles is the same as that in \figref{figure:weight}.
This motion consists of 8 postures:
1. arm down,
\, 2. elbow flexion,
\, 3. forearm pronation,
\, 4. forearm supination,
\, 5. raise upper arm, move hand forward,
\, 6. raise upper arm further, pull hand backward,
\, 7. lower upper arm backward,
\, 8. bring hand to the front of the chest.
Firstly, the pronation and supination of the forearm are performed 
using the branched muscle, which is placed at the position of the triceps brachii.
It achieves a rotation of $\pm$30 degrees.
Secondly, in the motion of raising the upper arm,
the two degrees of freedom of the deltoid 
are utilized to move the upper arm forward, middle, and backward.
While the muscle has a round shape that covers the joint,
it can function as a muscle.
After that, the motion of bringing the hand to the front of the chest 
is performed using muscle 2, the pectoralis major.
The motion is performed with a small current due to the presence of two wires.
Due to the interaction among muscles, 
each muscle undergoes deformation in bending or stretching 
in response to the contraction of other muscles. 
Even in such situations, 
PSMs maintain their own structure and deform and operate 
while preserving the wire path.
This contributes to the wide range of motion of the hand from the back to the front.
Through this experiment, it has been demonstrated that PSM 
can be used as muscle in an actual musculoskeletal robot.
Its capability to be created in various shapes and multi-DOF makes it useful
in the construction of musculoskeletal robots.
}
{%

次に、上腕全体を使った約30秒間の動作を行わせた。
これを\figref{figure:arm_merge}に示す。
図の上部にはその様子を示し、下部には各筋の電流およびワイヤ長さの時間変化を示す。
単筋であるMuscle 1および、\figref{figure:arm}で述べたMuscle 5のプロットは重ねて描画し、
残りは\figref{figure:weight}と同様に描画した。
この動作は、
1. 腕をおろした状態
\, 2. 肘を屈曲
\, 3. 肘の回外
\, 4. 肘の回内
\, 5. 上腕を上げ、手を前に出す
\, 6. 上腕をさらに上げ、後ろに引く
\, 7. 上腕を後方に下げる
\, 8. 手を胸の前に持ってくる
\,
の8つの姿勢を一連で行うものである。
まず、肘の回外および回内では、上腕三頭筋部分に配置した、二股に分かれた筋を活用し、
$\pm$30度の回転を実現している。
次に、上腕を上げる動作では、
三角筋の2自由度を活用し、上腕を前・中・後方へとそれぞれ動かすことに成功している。
また、関節を覆う形状でありながらも、筋肉としての機能が実現できていることが確認された。
最後に大胸筋を利用して、手を胸の前に持ってくる動作を行っているが、
ワイヤが2本あること、肩に関するモーメントアームが大きいことにより、
少しの電流で動作を実現できている。
筋同士が相互作用するため、各筋には曲げや伸びの変形が生じるが、
そのような状況でも筋自身の構造、ワイヤ経路を維持しながら変形、動作する。
よって、手が後方から前方まで、広い可動域で動作できていることもわかる。
この実験により、PSMが実際のロボットにおいてアクチュエータとして使用可能であることが示された。
また、様々な形状・多自由度で作成できることが、
筋骨格ロボットの構成において有用であることが示された。
}
\switchlanguage
{%

Furthermore, we performed simplified movements, 
such as raising the upper arm and bringing the hands in front of the chest, 
in situations involving environmental contacts. 
This is illustrated in \figref{figure:arm_env}.
Under the first two conditions,
the upper arm was moved with two types of bags put on the shoulder, 
containing 1.38kg weights.
The bags were placed to make contact with the pattern parts of the deltoid,
the pectoralis major mainly.
In particular, 75\% of the contact area of the bag on the shoulder
was directly in contact with the pattern part of the deltoid,
and the remaining 25\% was in contact with the end point of the deltoid.
In such situations where the muscles are loaded by environmental contacts,
PSM is able to operate, allowing the upper arm to move.
Also, in the remaining two conditions,
the upper arm was moved with cloth wrapped around the structure,
and with a person directly touching the pattern part of the deltoid.
From the third experimental condition,
it was confirmed that the motion could be performed without catching the cloth 
at their muscles and joints.
From the final experimental condition,
it was shown that safety is ensured by covering joints and wires 
with muscles composed of flexible materials, 
and even when the muscles are touched directly by a person,
the operation can be performed.
In summary, using PSM as muscle in a musculoskeletal robot 
allows for safe movements in various environments, 
involving environmental contacts.
}
{%

さらに、より動作を簡略化し、上腕を上げる、胸の前に手を持ってくる動作を、
複数の環境接触を伴う状況で行った。これを\figref{figure:arm_env}に示す。
最初の2つの条件では、2種類の、1.38kgの重りが入ったバッグを肩にかけ、
三角筋、大胸筋等のパターン部分にバッグが接触する状況で動作させた。
特に、バッグの肩への接触部分の75\%は、三角筋のパターン部分に直接接して荷重をかけており、
残りの25\%は骨との接合部に接する。
このような、環境接触により筋に荷重がかかる状況でも、PSMは問題なく動作し、
上腕を動かすことができた。
また、残りの2つの条件では、上腕構造に布を巻いた状態および、
人が三角筋のパターン部分を直接触る状況で動作させた。
布を巻き込むことのない動作ができること、
柔軟素材で関節やワイヤを覆っているために安全で、
動作中に筋を直接触っても問題ないことが確認された。
}%

\section{Conclusion}\label{sec:conclusion}
\switchlanguage
{
In this study, we proposed Patterned Structure Muscle (PSM), which can be used as a muscle for musculoskeletal robots.
In the design of PSM, we utilized the anisotropic characteristics of the pattern structure,
FDM 3D printing of flexible material TPU, and wire-driven actuation.
This approach allows for the movement of muscles with various shapes and degrees of freedom 
under environmental contact forces.
In the experiment, 
we confirmed the lifting of heavy objects and operation under environmental contact in various shapes of PSMs. 
We also confirmed that in the upper arm structure composed of multiple PSMs, 
strong movements, a wide range of motion, and environmental contact operations are achieved.
All of these results demonstrate that PSM is suitable for use as muscles in musculoskeletal robots.
%

PSM is highly scalable because of its sparse-structure and 3D printing making process. 
By using conductive filaments to detect deformation or embedding sensors, 
it is believed that various functionalities, including tactile sensing, will be achieved. 
Also, in the future, we will create a musculoskeletal humanoid with a structure that mimics the human body using PSM.
We think that PSM will enable body movements closer to those of human,
because PSM can imitate the complex movements of vairous human muscles.
Furthermore, by utilizing environmental contact performance and safety, 
it is believed that significant interactions with external environments and humans, 
such as rolling over motiom or hugging, will be realized.
}
{
本研究では、筋骨格ロボットの筋として利用可能な構造として、
PSMを提案した。
この構造の作成においては、パターン構造を活用した異方的な特性、柔軟材料TPUの積層3Dプリント、
ワイヤ駆動の利用を組み合わせた。
よって、様々な形状・多自由度で、環境からの接触力のもとでも、
筋の内部構造の位置関係を維持しながらの運動が可能である。
実際に、パラメータを変えたPSMの特性を計測し、
この構造の異方性な特性を確認し、
パラメータを変えることでの性能の調整も可能であることを示した。
次に、PSMの単体動作を確認し、重量物持ち上げ、環境接触下での動作ができることを確認した。
多自由度・様々な形状の筋においても、その内部構造の位置関係を維持しながら動作ができることを示した。
また、収縮力はモータの性能に依存するため、モータの性能に応じて大きな引張力を発揮可能であった。
最後に、複数のPSMで構成した上腕構造について、外部物体や人との接触を伴う複数環境での動作を行った。
多自由度の筋や、関節を覆う筋が、実際の筋骨格ロボットにおいて
アクチュエータとして使用可能であることが示された。

今後の展望としては、このPSMを活用し、
より人体を模倣した構造の筋骨格ロボットを作成することや、
直列弾性をはじめとする、人間の筋が有する特性の追加
が挙げられる。

このPSMは、疎構造の3Dプリントであるために拡張性が高い。
導電性フィラメントを使用し変形を検知したり、センサを埋め込むことで、触覚をはじめとする
多様な機能を持たせることができると考えられる。
また、将来、このPSMを活用して、人体模倣の筋骨格ロボットを作成すると、
筋骨格構造が人間により近く、人間により近い動きを実現できると考えられる。
さらに、環境接触性能や安全性を活用することで、寝返りや抱きつきなどの、外部環境や人との大きなインタラクションが必要な
動作も実現可能になると考えられる。
}%

{
  \bibliographystyle{IEEEtran}
  \bibliography{bib}
}

\end{document}